\documentclass[10pt,twocolumn,letterpaper]{article}

\usepackage[pagenumbers]{cvpr} 

\usepackage[dvipsnames]{xcolor}

\usepackage{enumitem}
\usepackage{booktabs}

\usepackage{times}
\usepackage{epsfig}
\usepackage{graphicx}
\usepackage{amsmath}
\usepackage{amssymb}

\makeatletter
\DeclareRobustCommand\onedot{\futurelet\@let@token\@onedot}
\def\@onedot{\ifx\@let@token.\else.\null\fi\xspace}

\usepackage{multirow}
\usepackage{algorithm}
\usepackage[noend]{algorithmic}
\usepackage{graphicx, amsmath, amssymb, caption, subcaption, multirow, overpic, textpos}

\def\eg{\emph{e.g}\onedot} 
\def\ie{\emph{i.e}\onedot} 
 
 \def\vs{\emph{vs}\onedot}

\newcommand{\figref}[1]{Fig\onedot~\ref{#1}}

\newcommand{\secref}[1]{Sec\onedot~\ref{#1}}
\newcommand{\tabref}[1]{Tab\onedot~\ref{#1}}

\newlength\secmargin
\newlength\paramargin
\newlength\abovetabcapmargin
\newlength\belowtabcapmargin
\newlength\abovefigcapmargin
\newlength\belowfigcapmargin

\usepackage{pifont}

\usepackage{xcolor}

\definecolor{datasegcolor}{gray}{.8}

\definecolor{baselinecolor}{gray}{.9}

\newlength\savewidth\newcommand\shline{\noalign{\global\savewidth\arrayrulewidth
  \global\arrayrulewidth 1pt}\hline\noalign{\global\arrayrulewidth\savewidth}}
\newcommand{\tablestyle}[2]{\setlength{\tabcolsep}{#1}\renewcommand{\arraystretch}{#2}\centering\footnotesize}

\definecolor{cvprblue}{rgb}{0.21,0.49,0.74}
\usepackage[pagebackref,breaklinks,colorlinks,citecolor=cvprblue]{hyperref}

\usepackage[capitalize]{cleveref}
\crefname{section}{Sec.}{Secs.}
\Crefname{section}{Section}{Sections}
\Crefname{table}{Table}{Tables}
\crefname{table}{Tab.}{Tabs.}
\def\paperID{***} 
\def\confName{CVPR}
\def\confYear{2024}

\newcommand{\modelnamefull}{OmniScient Model\xspace}
\newcommand{\modelname}{OSM\xspace}
\newcommand{\bridgename}{MaskQ-Former\xspace}
\newcommand{\queryname}{Mode Query\xspace}

\title{Towards Open-Ended Visual Recognition with Large Language Model
}

\author{Qihang Yu\quad\quad Xiaohui Shen\quad\quad Liang-Chieh Chen\\
ByteDance
}

\begin{document}
\maketitle
\begin{abstract}
Localizing and recognizing objects in the open-ended physical world poses a long-standing challenge within the domain of machine perception.
Recent methods have endeavored to address the issue by employing a class-agnostic mask (or box) proposal model, complemented by an open-vocabulary classifier (\eg, CLIP) using pre-extracted text embeddings. However, it is worth noting that these open-vocabulary recognition models still exhibit limitations in practical applications.
On one hand, they rely on the provision of class names during testing, where the recognition performance heavily depends on this predefined set of semantic classes by users.
On the other hand, when training with multiple datasets, human intervention is required to alleviate the label definition conflict between them. 
In this paper, we introduce the \modelnamefull (\modelname), a novel Large Language Model (LLM) based mask classifier, as a straightforward and effective solution to the aforementioned challenges. 
Specifically, \modelname predicts class labels in a generative manner, thus removing the supply of class names during both training and testing.
It also enables cross-dataset training without any human interference, exhibiting robust generalization capabilities due to the world knowledge acquired from the LLM.
By combining \modelname with an off-the-shelf mask proposal model, we present promising results on various benchmarks, and demonstrate its effectiveness in handling novel concepts.
Code/model are available at \url{https://github.com/bytedance/OmniScient-Model}.
\end{abstract}    
\vspace{-0.5ex}
\section{Introduction}
\vspace{-0.5ex}
\label{sec:intro}

\begin{figure}[th]
    \centering
    \includegraphics[width=0.8\linewidth]{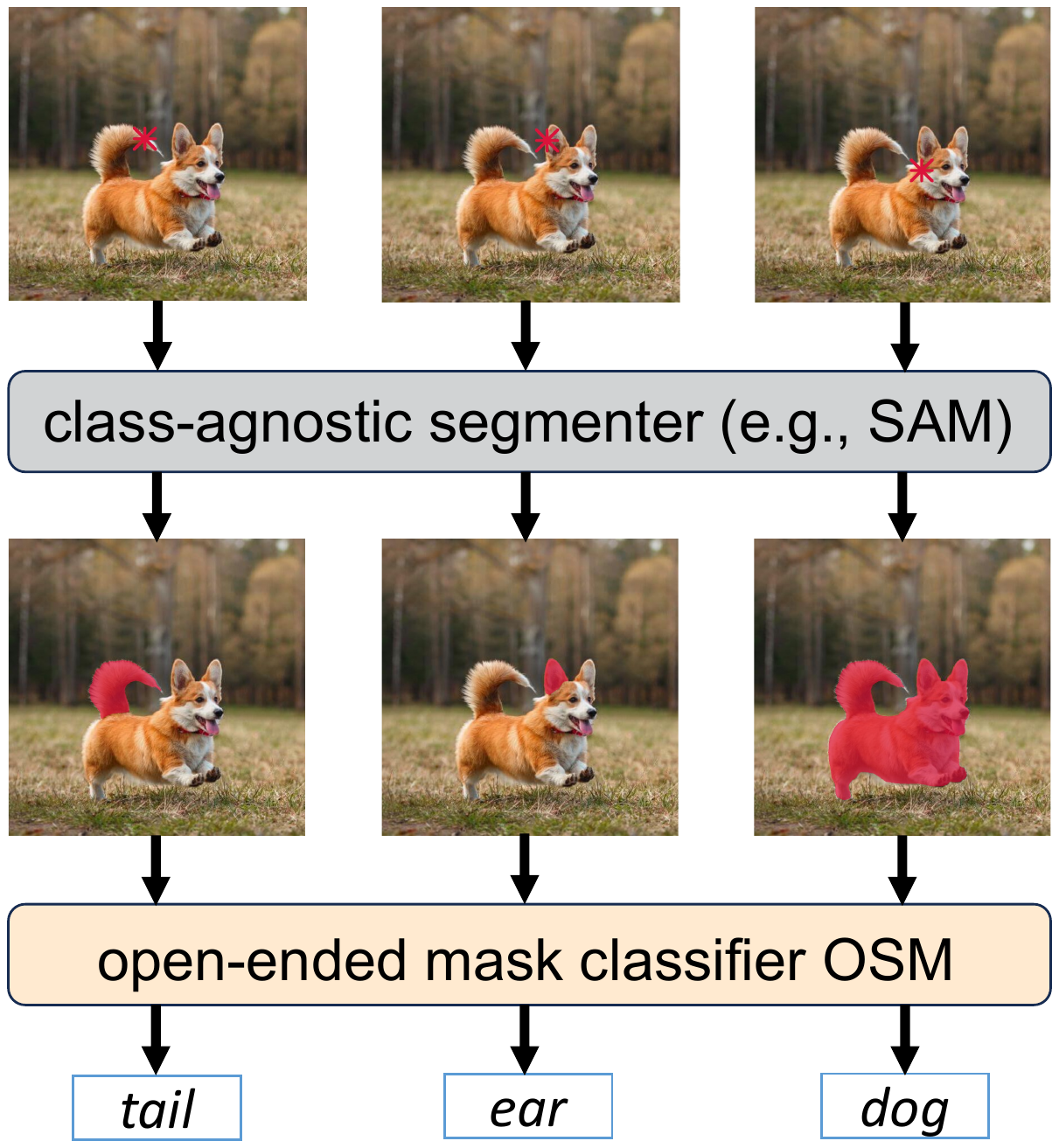}
    \vspace{-6ex}
    \caption{
    \textbf{Illustration of open-ended recognition.}
    The open-ended recognition task is decomposed into two sub-tasks: class-agnostic mask proposal and open-ended mask classification. To tackle the task, we propose \modelname (\modelnamefull), an open-ended recognition model that works hand in hand with an off-the-shelf class-agnostic mask proposal model (\eg, SAM). Unlike existing open-vocabulary recognition models, \modelname does not require any user-predefined vocabulary and instead directly predicts the class of each proposal with unconstrained vocabulary in a generative manner. As a result, \modelname shows a great generalization ability. For example, we observe \textbf{emergent part predictions} such as tail and ear, while \modelname has never seen such masks or labels during training (\ie, we do not use any part segmentation datasets during training). Moreover, by obtaining masks from a class-agnostic segmenter, we can take advantage of it and take a wide range of prompt types including point, box, and mask.
    }
    \vspace{-3ex}
    \label{fig:teaser_img}
\end{figure}

A persistent challenge in the realm of machine perception involves the accurate localization~\cite{girshick2015fast,long2015fully,chen015deeplabv1,he2017mask,chen2017deeplab} and recognition~\cite{deng2009imagenet,krizhevsky2012imagenet,he2016deep} of objects in real-world settings.
While considerable progress has been made on various standard benchmarks~\cite{radford2021learning,jia2021scaling,yu2022coca,wang2023internimage}, existing methods continue to grapple with the complexities of real-life scenarios where novel out-of-training dataset concepts frequently arise.
To address this issue and enhance the practical utility of models, a common strategy is to decompose the problem into two components: class-agnostic mask/box proposal and mask/box classification, as highlighted in previous works~\cite{carion2020end,zhu2020deformable, wang2021max,cheng2021per}.
It has been observed that mask/box proposal models, when trained on a dataset such as COCO~\cite{lin2014microsoft}, can still effectively generalize to previously unseen concepts~\cite{zhou2022detecting,kim2022learning}. 
Additionally, recent advancements, exemplified by Segment Anything Model (SAM)~\cite{kirillov2023sa-1b}, have expanded the training dataset to an extensive scale, encompassing 1.1 billion class-agnostic masks from 11 million images.
This has yielded a mask proposal model characterized by robust zero-shot segmentation capabilities, generalizing to novel images and concepts.
These developments present a promising avenue toward a solution to the first issue regarding  class-agnostic object proposals.

Despite the remarkable achievements in the development of general proposal models, addressing the challenge of classifying novel concepts in real-world scenarios remains an unsolved issue. Many of the existing approaches leverage vision-language models (VLMs), such as CLIP~\cite{radford2021learning} and ALIGN~\cite{jia2021scaling}, which have been pretrained on extensive Internet datasets and have demonstrated outstanding performance in aligning images and text within a shared embedding space. Specifically, these techniques~\cite{ghiasi2022scaling,zhou2022detecting,ding2022open,yu2023convolutions,liang2023open,xu2023open,wang2023detecting} aim to train open-vocabulary classifiers that rely on the precomputed text embeddings derived from VLMs, as opposed to learning label embeddings directly from the training dataset. The dependency on VLM text embeddings highlights the inherent power and generalization capabilities of VLMs, which, to a certain extent, ensure the classifier's ability to generalize to novel concepts.

Nevertheless, it is important to acknowledge that while these methods have shown promise, they are still confronted with several challenges that impede their practical application. Firstly, these models typically operate under the assumption that class names are predefined during testing, a condition seldom met in real-life scenarios.
Furthermore, when utilizing multiple diverse datasets, complications arise when different label definitions or label space conflicts exist among them.
Consequently, many current multi-dataset frameworks address this issue by training on each dataset with an individual decoder or classifier~\cite{zhou2022simple,gu2023dataseg,zhou2023lmseg}, or merge the label space manually~\cite{lambert2020mseg}, adding complexity to the process.

To address these challenges, we introduce \modelnamefull (\modelname), a novel generative framework towards open-ended recognition tasks. Instead of training the model to ``select" correct classes from a predefined vocabulary, our approach focuses on training it to generate the desired class names.
This paradigm shift means that the model no longer requires the prior knowledge of all possible class names provided by users, eliminating the necessity for a well-defined vocabulary during both training and testing phases.
Consequently, this approach naturally accommodates the training and testing on datasets with varying label spaces, obviating the need for human intervention to harmonize the differences.
Additionally, by building upon a pre-trained Large Language Model (LLM)~\cite{touvron2023llama,vicuna2023}, \modelname leverages the implicitly learned world knowledge~\cite{roberts2020much,hendrycks2020measuring} encoded within the LLM, enhancing its ability to effectively generalize to novel concepts, further bolstering its utility and reliability.

We conduct meticulous experiments to validate the appropriateness of employing a generative model for discriminative tasks. Our investigation includes assessing the generative model's ability to effectively capture and adapt to the characteristics of a given training dataset and its associated vocabulary. We compare its performance to that of a discriminative model, primarily focusing on classification accuracy.
Additionally, we introduce a \queryname mechanism, which empowers the model to make predictions within a predefined vocabulary (referred to as vocabulary-specific predictions), or to provide open-ended predictions without vocabulary constraints (referred to as vocabulary-agnostic predictions).
Finally, we integrate \modelname with various off-the-shelf segmentors (\ie, mask proposal models), such as kMaX-DeepLab~\cite{yu2022k} and SAM~\cite{kirillov2023sa-1b}, and validate its effectiveness across several benchmarks.

\vspace{-0.5ex}
\section{Related Work}
\vspace{-0.5ex}
\label{sec:related_work}

\quad\textbf{Open-Vocabulary Recognition}\quad
Recently, exemplified by CLIP~\cite{radford2021learning} and ALIGN~\cite{jia2021scaling}, open-vocabulary recognition methods have demonstrated promising outcomes. These methods involve the pre-training of dual-encoder models (for image and text) using contrastive objectives on extensive collections of noisy image-text pairs. This pre-training process yields feature representations that possess cross-model capabilities, showcasing robust performance in zero-shot downstream tasks. Drawing inspiration from these advances, the field of open-vocabulary detection and segmentation has also witnessed remarkable breakthroughs, where class names provided during testing may not have been encountered during the training phase.
A majority of these state-of-the-art techniques~\cite{ghiasi2022scaling, zhou2022detecting, ding2022open, yu2023convolutions, liang2023open, wang2023detecting} approach the problem by disentangling it into class-agnostic proposals, along with open-vocabulary proposal classification by leveraging a pre-trained CLIP model. However, despite the impressive accomplishments of these open-vocabulary methods in recognizing unseen classes beyond the training dataset, they hinge on a strong yet brittle assumption that the semantic classes (\ie, vocabulary) are known in advance and remain static, an assumption that can easily be disrupted in practical applications.
In parallel with our research efforts, vocabulary-free image classification~\cite{conti2023vocabulary} seeks to address this challenge by dynamically generating vocabularies through processes such as parsing captions~\cite{li2022blip,li2023blip} or retrieving them from external databases~\cite{schuhmann2022laion}.
By contrast, our approach offers a straightforward solution by reformulating the open-ended classification problem as text generation~\cite{sutskever2014sequence,devlin2018bert,radford2019language,brown2020language},  naturally eliminating the need for a user-predefined vocabulary.
Furthermore, while our method primarily focuses on object-level recognition, the work presented in~\cite{conti2023vocabulary} concentrates on image-level classification.

\textbf{Large Language Models}\quad
In recent years, research community has witnessed a remarkable surge in the development of Large Language Models (LLMs)~\cite{brown2020language, openai2023gpt, thoppilan2022lamda, raffel2020exploring, touvron2023llama}.
These models have demonstrated impressive emergent capabilities, including in-context learning~\cite{brown2020language}, instruction following~\cite{wei2021finetuned, chung2022scaling}, and chain-of-thought reasoning~\cite{wei2022chain}.
However, a significant limitation of these LLMs is their inherent ``blindness" to other modalities, such as visual inputs.
More recently, the excitement surrounding multi-modal LLMs has surged, particularly with the introduction of GPT-4V~\cite{openai2023gpt}. Pioneering research~\cite{li2023blip, zhu2023minigpt,liu2023visual,liu2023improved, alayrac2022flamingo,chen2023shikra,wang2023visionllm} has illustrated a promising avenue for bridging the gap between language and vision modalities. This approach involves constructing modular models that typically consist of a frozen CLIP vision encoder, a trainable bridging module (\eg, Perceiver Resampler in~\cite{alayrac2022flamingo}, Q-Former in~\cite{li2023blip}, or a simple linear/MLP layer in~\cite{liu2023visual, liu2023improved}), and a frozen LLM. Furthermore, \cite{zhao2023bubogpt,wang2023visionllm,peng2023kosmos,chen2023shikra,zhang2023gpt4roi} add referring or grounding ability to the multi-modal LLM through taking bounding-box as inputs or outputs.
The proposed \modelname can be categorized as a modular multi-modal LLMs with referring capability.
However, previous endeavors primarily aim to enhance multi-modal LLMs with bounding-box (as bounding-box can be naturally represented in text by referring to its coordinates) for conversation applications, which also require providing vocabulary in the input prompt~\cite{wang2023visionllm,zhang2023gpt4roi}.
Our perspective underscores the value of enabling multi-modal LLMs to recognize \textit{segmentation masks} and serve as standalone tools.
\section{Method}
\label{sec:method}

In this section, we introduce \modelname (\modelnamefull), an open-ended recognizer.
We commence by detailing how we transform the conventional classification task into a text generation task, aligning with the principles outlined in~\cite{radford2019language,brown2020language} (\secref{sec:problem}).
Subsequently, we elucidate the construction of \modelname, which follows the footsteps of previous modular vision-language models~\cite{li2023blip,instructblip, liu2023visual,zhu2023minigpt} (\secref{sec:arch}).
We also provide a comprehensive overview of our training and evaluation protocols (\secref{sec:protocols}).

\subsection{Problem Formulation of Classification}
\label{sec:problem}
Without loss of generality, we focus our discussion on mask classification.
Given an input image $\mathbf{I} \in \mathbb{R}^{H \times W \times 3}$ and a collection of $M$ segmentation masks $\mathbf{M} \in \mathbb{R}^{H \times W \times M}$ (from an off-the-shelf segmenter, \eg, SAM~\cite{kirillov2023sa-1b}), our objective is to predict a semantic class for each of these masks:
\begin{equation}
\{y_i\}_{i=1}^M = \{(m_i, c_i)\}_{i=1}^M ,
\end{equation}
where $m_i$ is the $i$-th mask from $\mathbf{M}$ and $c_i$ is its predicted class,  belonging to the set of predefined semantic classes $C$, which is assumed to be known during both training and testing phases.
In a closed-vocabulary setting, models focus solely on the target classes, implying that the set of predefined semantic classes are identical during both training and testing (\ie, $C_{train} = C_{test}$, where the subscript denotes the training or testing phase).
By contrast, in an open-vocabulary setting, this assumption is relaxed by allowing for the possibility that $C_{test}$ may include novel categories that were not seen during training (\ie, $C_{test}\neq C_{train}$). Nevertheless, in both cases, it is essential to have access to the category names of $C_{train}$ and $C_{test}$ during both the training and testing stages.
As a result, the recognition performance heavily hinges on the careful design of $C_{train}$ and $C_{test}$ (called prompt engineering in the literature~\cite{ghiasi2022scaling,yu2023convolutions}).

\begin{figure*}[t!]
    \centering
    \vspace{-22ex}
    \includegraphics[width=0.8\linewidth]{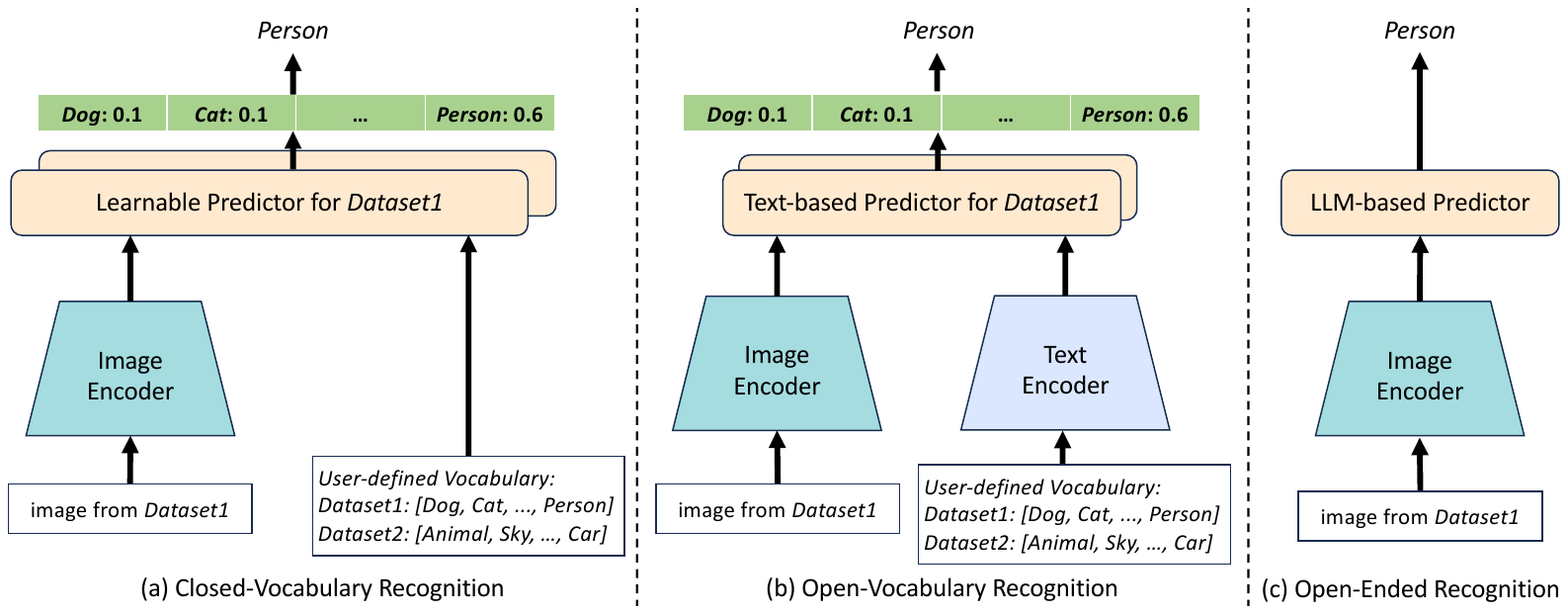}
    \vspace{-19ex}
    \caption{
    \textbf{Comparisons between recognition schemes.}
    (a) In the closed-vocabulary recognition setting, the sets of semantic classes are fixed during both training and testing.
    A learnable predictor (\eg, $1\times 1$ convolution layer) is used for each training dataset.
    (b) In the open-vocabulary recognition setting, the sets of semantic classes can be different during training and testing, allowing detection of novel concepts during testing by leveraging a pretrained CLIP backbone.
    The text-based predictor (\ie, the text embeddings of the predefined set of semantic classes) is different for each dataset.
    (c) In the open-ended recognition setting, the model directly predicts the class names in a generative manner, removing the need to predefine the semantic classes during training and testing. Additionally, it enables the cross-dataset training in an easier way (\eg, no need to involve humans to resolve the label definition conflicts between datasets).
    }
    \label{fig:task_comp}
\end{figure*}

The aforementioned assumption (\ie, the access to $C_{train}$ and $C_{test}$) plays a pivotal role in contemporary recognition frameworks, whether operating in a closed-vocabulary or open-vocabulary context. These frameworks typically rely on computing similarity logits across semantic class candidates and selecting the candidate with the highest probability as the final prediction. While these methods have demonstrated effectiveness and success across various tasks and benchmarks over the past decades, they are not without critical limitations.
Firstly, it is practically impossible to predefine and encompass all potential semantic classes present in the real world. This limitation poses a significant challenge in open-vocabulary recognition since it necessitates the prior definition of novel concepts within the vocabulary. Furthermore, many of these methods are constructed around a handcrafted and meticulously designed label space, with the expectation of covering common concepts that should ideally have unambiguous definitions. However, the manual curation of label spaces may not be scalable, particularly when researchers aim to expand their models to encompass all available datasets from various sources. This process may require labor-intensive tasks such as meticulous manual merging~\cite{lambert2020mseg} or conducting separate training~\cite{gu2023dataseg, zhou2023lmseg}.

To address those challenges, we depart from the conventional approach in visual recognition and propose a novel paradigm named open-ended visual recognition. In this paradigm, we make the bold assumption that the vocabulary $C$ remains \textbf{unknown} during both training and testing.
We note that during training we only need to access the target class for each mask, without the need to know the existence of all the other possible classes in $C$, which is required for existing methods relying on softmax-based prediction.
This shift in perspective is illustrated in~\figref{fig:task_comp} for a holistic comparison of the different paradigms.
Rather than selecting a prediction class from a predefined vocabulary, our approach involves directly predicting the class name of the target object. Essentially, this reformulates the recognition task as a text generation problem. Mathematically, we frame open-ended recognition as an endeavor to maximize the conditional likelihood of the class name under a forward autoregressive factorization:
\begin{equation}
p(c_{i}) = \prod \limits_{j=0}^N p(c_{i,j}|c_{i,0},\cdots,c_{i,j-1}),
\end{equation}
where $c_{i,j}$ corresponds to the $j$-th text token within the class names for $c_{i}$.

\subsection{Model Architecture}
\label{sec:arch}

\definecolor{myblue}{rgb}{0.69019608,0.85882353,0.8745098}
\definecolor{myyellow}{rgb}{0.98431373,0.92156863,0.82745098}
\definecolor{mygreen}{rgb}{0.89803922,0.94117647,0.85882353}
\begin{figure*}[t]
    \centering
    \vspace{-20ex}
    \includegraphics[width=0.85\linewidth]{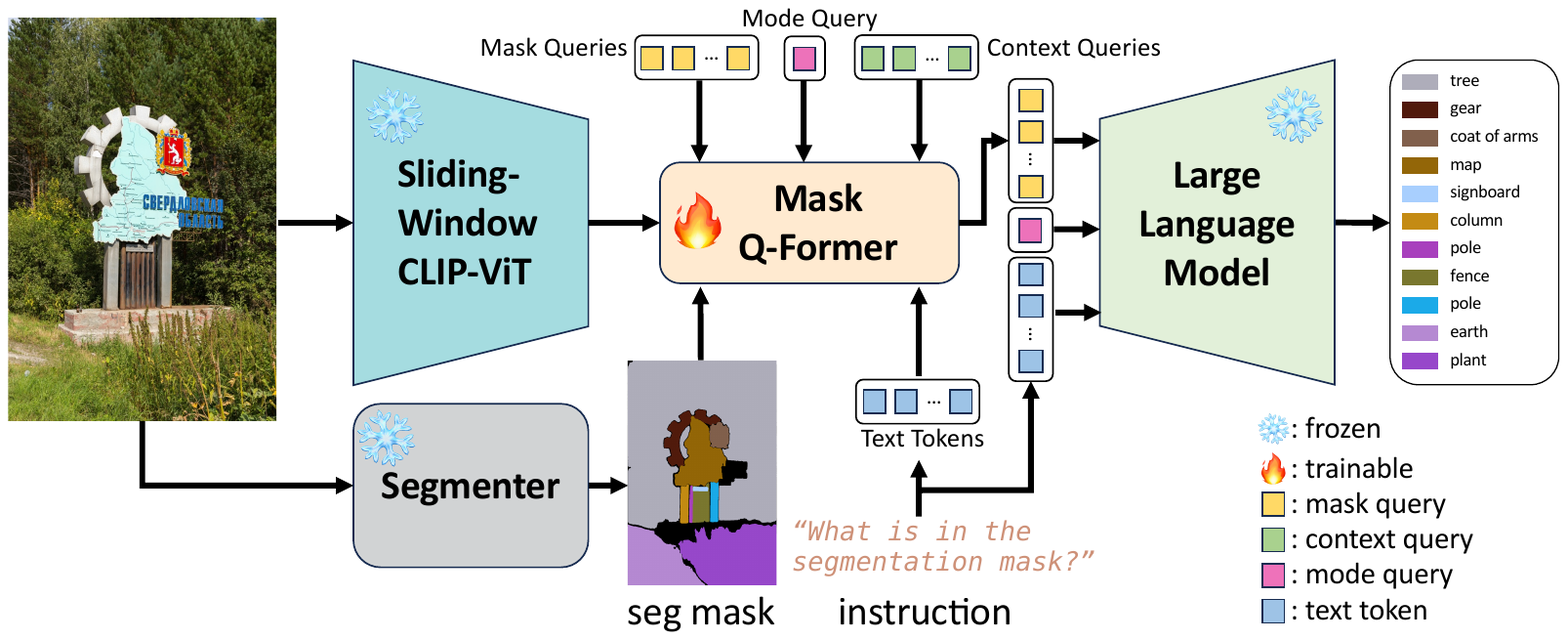}
    \vspace{-18ex}
    \caption{\textbf{An overview of the proposed \modelname,} consisting of a frozen \textcolor{NavyBlue}{CLIP-ViT} that extracts high-resolution features in a sliding-window manner, a trainable \textcolor{Peach}{\bridgename} that resamples pixel features in a mask-aware manner, and a frozen \textcolor{ForestGreen}{LLM}, which predicts a semantic class for each corresponding mask in a generative manner without a predefined vocabulary.
    \modelname can be combined with any off-she-shelf segmenter, \eg, SAM~\cite{kirillov2023sa-1b} and kMaX-DeepLab~\cite{yu2022k}.
    The proposed \bridgename takes as input (1) Mask Queries, (2) Context Queries, and (3) \queryname.
    The Mask Queries focus on the mask regions proposed by the off-the-shelf segmenter, while the Context Queries attend to a broader region derived from the mask.
    The \queryname consists of two modes: vocabulary-specific and vocabulary-agnostic, allowing the model to perform with dataset-specific and dataset-agnostic vocabularies, respectively. Note that we have two separate Model Query for \bridgename and LLM respectively, and only the Mask Queries from MaskQ-Former are fed into LLM.
    }
    \label{fig:mode_arch}
\end{figure*}

The architectural overview of \modelname is presented in~\figref{fig:mode_arch}. In alignment with the established modular vision-language models~\cite{li2023blip,zhu2023minigpt,liu2023visual}, \modelname comprises three principal components: a frozen CLIP-ViT, a trainable \bridgename, and a frozen Large Language Model (LLM). Our approach incorporates specialized design enhancements aimed at optimizing the model for object-level recognition, which we detail in the following paragraphs:

\vspace{0.5ex}
\textbf{High-Resolution Feature Extraction with Frozen CLIP-ViT}\quad
A frozen vision transformer backbone, pre-trained in the CLIP style, has become the standard choice in existing multi-modal LLM designs. The appeal of CLIP-ViT lies in its dual advantages: it provides a robust and adaptable feature representation for input images, and its feature space is well-suited for seamless conversion into language tokens, which the LLM can comprehend as inputs.

Nonetheless, the usage of CLIP-ViT, while successful in many multi-modal LLM applications such as image captioning~\cite{chen2015microsoft,plummer2015flickr30k} and visual question answering~\cite{goyal2017making,hudson2019gqa}, has its limitations. It was originally pre-trained on lower resolutions, typically at resolution $224\times224$. This lower resolution can hinder its performance, especially when tasked with object-level recognition. Moreover, previous research~\cite{yu2023convolutions} has observed that a frozen ViT exhibits weak generalization capabilities across varying input resolutions.

Despite the widespread use of frozen ViT backbones in multi-modal LLM models, it is evident that a $224\times224$ input resolution falls short, particularly for object-level recognition. Typical adaptations, such as windowed attention~\cite{liu2021swin} as seen in ViTDet~\cite{li2022exploring}, may not be applicable to a completely frozen ViT backbone. To address this limitation, we propose a straightforward strategy to extract more effective features using a frozen ViT at a higher resolution, for example, $896\times 896$. Specifically, we employ a sliding-window feature extraction approach at the input level, where each window size matches that of the ViT's pre-trained image size. Afterwards, a global positional embedding is added to compensate the missing location information across windows.
In our experiments, we will empirically prove that this seemingly simple strategy is surprisingly effective, yielding significantly improved performance in feature extraction from high-resolution inputs.
\textbf{\bridgename}\quad
We employ a visual resampler, such as Q-Former~\cite{li2023blip} or Perceiver Resampler~\cite{alayrac2022flamingo}, to bridge the gap between the encoded image features and inputs suitable for the LLM. This visual resampler typically consists of a stack of transformer decoders~\cite{vaswani2017attention} that transform image tokens into a reduced set of query tokens, which are usually far fewer in number compared to image tokens. However, existing visual resamplers like those used in~\cite{li2023blip, alayrac2022flamingo}, employ a set of queries that globally attend to image features without considering the segmentation mask priors.

\begin{figure}[t!]
    \centering
    \vspace{-5ex}
    \includegraphics[width=1.0\linewidth]{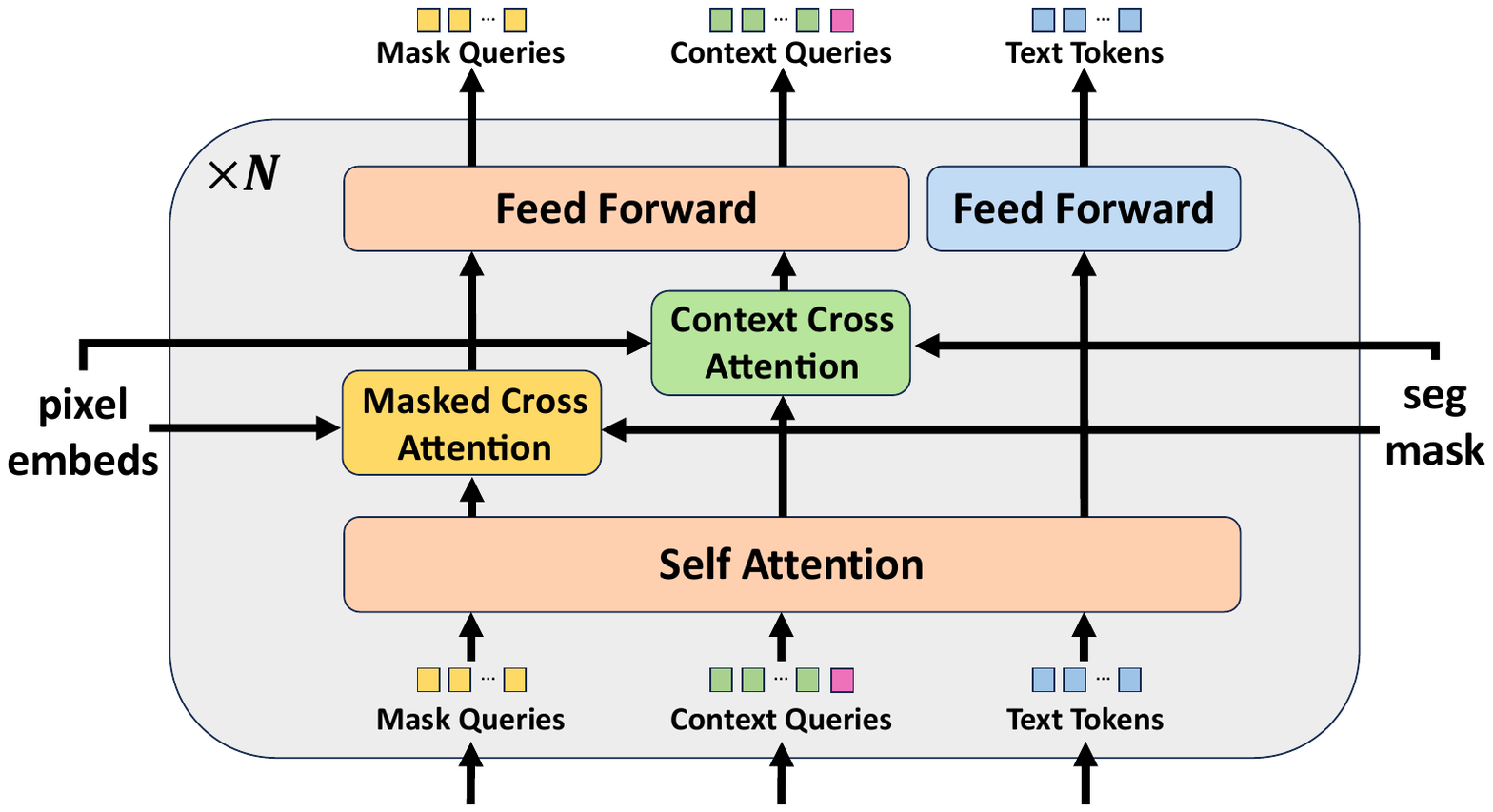}
    \vspace{-10ex}
    \caption{\textbf{An overview of \bridgename.} The parameters of Masked Cross Attention layer and Context Cross Attention layer are shared. We append the \textcolor{purple}{Mode Query} to \textcolor{YellowGreen}{Context Queries}. Moreover, \textcolor{Dandelion}{Mask Queries} only attend to the mask region in cross-attention layer, while \textcolor{YellowGreen}{Context Queries} may attend to a larger region derived from the mask. All queries/tokens will communicate with each other in the self-attention layer.
    }
    \label{fig:mcqformer_arch}
\end{figure}

In response to this limitation, we introduce a novel variant called \bridgename. The \bridgename takes a segmentation mask as input and performs masked cross-attention~\cite{cheng2022masked}, as depicted in~\figref{fig:mcqformer_arch}. It consists of two sets of learnable queries: mask queries and context queries. The mask queries execute masked cross-attention, restricting their focus to the mask region, while the context queries attend to a broader region derived from the mask, such as the bounding box region, to provide complementary contextual information. This contextual information is essential for precise and unbiased recognition~\cite{chen2017rethinking,chen2018encoder,yu2020context}.

The \bridgename effectively summarizes the mask region while retaining access to essential contextual content. Information exchange between the mask queries and context queries is facilitated through the self-attention layer. Notably, all parameters are shared between the mask and context queries, except for the learnable query initialization, resulting in negligible additional costs. In the end, we retain only the mask queries as inputs to the LLM, ensuring computational efficiency.

\textbf{\queryname}\quad
While our primary objective is to enable \modelname to perform effectively in an open-ended setting, where it can make predictions without prior knowledge of any vocabulary, we acknowledge the importance of versatility. \modelname has the capability to perform accurately when required to align with a specific vocabulary. To achieve this, we introduce \queryname, consisting of vocabulary-specific and vocabulary-agnostic queries, drawing inspiration from prefix tuning techniques~\cite{li2021prefix}. These queries leverage the strong instruction-following capabilities of the LLM, enhancing the model's adaptability across diverse scenarios.
Concretely, we propose appending a dedicated learnable query for each vocabulary to both the \bridgename and LLM inputs. During training, when utilizing datasets from various sources, the corresponding vocabulary-specific query for each dataset is activated, allowing the model to effectively ``memorize" the associated vocabulary of each dataset, thereby improving alignment during prediction. Additionally, we include a general vocabulary-agnostic query that is activated during training on any dataset to keep the open-ended recognition ability.

This approach provides flexibility during testing. We can activate a vocabulary-specific query to ensure that the model's predictions align better with the desired vocabulary, or we can activate the vocabulary-agnostic query to facilitate open-ended predictions. This adaptability enhances \modelname's utility across a spectrum of real-world scenarios, making it a versatile tool for a wide range of applications.

\subsection{Training and Evaluation Protocols}
\label{sec:protocols}

\quad\textbf{Datasets}\quad
To create a robust training and evaluation framework, we ensemble six publicly available segmentation datasets, encompassing diverse image distributions, domains, and segmentation tasks. These datasets include COCO panoptic segmentation~\cite{lin2014microsoft}, ADE20K panoptic segmentation~\cite{zhou2017scene}, Cityscapes panoptic segmentation~\cite{cordts2016cityscapes}, LVIS instance segmentation~\cite{gupta2019lvis}, ADE-847 semantic segmentation~\cite{zhou2017scene}, and PC-459 semantic segmentation~\cite{everingham2010pascal}.

\textbf{Training Protocols}\quad
During training, we adopt an instruction tuning approach~\cite{wei2021finetuned, chung2022scaling, liu2023visual} to seamlessly integrate training with the LLM. For each training iteration, we randomly select an image and its corresponding ground-truth mask from a dataset. We randomly choose an instruction template and insert the actual class name. This approach enables training the model using a straightforward next-token prediction loss without the need for intricate designs. We default to the template \textit{What is in the segmentation mask?} and greedy search decoding for testing.

The choice of training batch size varies across datasets, with batch size $32$ for COCO, $64$ for LVIS, $16$ for ADE-847, $8$ for PC-459, $16$ for ADE20K, and $8$ for Cityscapes, respectively. In each training batch, half of the data activate vocabulary-specific queries corresponding to their respective datasets, while the other half activate vocabulary-agnostic queries. We use AdamW optimizer~\cite{kingma2014adam,loshchilov2017decoupled} with learning rate $4\times10^{-5}$ and weight decay $0.05$. The learning rate follows a cosine decay schedule. Training continues until the model has processed a total of $6$ million masks.

\textbf{Evaluation Protocols}\quad
Our model is evaluated on the validation set of each dataset, using two types of masks: ground-truth masks or masks produced by an off-the-shelf segmenter.
When using ground-truth masks as inputs, we purely assess mask classification accuracy.
Specifically, a prediction is considered correct only when the predicted class name \textbf{exactly} matches the class name in the ground-truth annotation.
To enhance the reliability of this metric, we augment the ground-truth class names with synonyms sourced from~\cite{ghiasi2022scaling,yu2023convolutions}.
Additionally, we consider plural and singular formats of class names.
It is important to note that these synonyms are not used during model training, as they may not always be semantically aligned (\eg, ``person", ``man", and ``woman" are synonyms in COCO and LVIS).
As a result, we report two metrics: Accuracy (Acc) and Not-in-Vocabulary (NIV), which represent the percentage of predictions correctly match ground-truth classes, or the predictions do not fall into the dataset's vocabulary, respectively.
The metric Acc directly evaluates the model's classification capacity, while NIV reflects the model's generalizability or degrees of overfitting to the trained datasets.

Additionally, we consider a more practical application where \modelname is connected to an off-the-shelf mask proposal model, such as kMaX-DeepLab~\cite{yu2022k} or SAM~\cite{kirillov2023sa-1b}.
We directly evaluate the model's performance on the established academic benchmarks, including panoptic segmentation and semantic segmentation, using panoptic quality (PQ)~\cite{kirillov2019panoptic} and mIoU~\cite{everingham2010pascal}, respectively.

\section{Experimental Results}
\label{sec:experiments}

In this section, we first provide the settings used for the ablation studies and our final model.
We then evaluate \modelname with ground-truth masks along with ablation studies in~\secref{sec:gt_masks}, followed by the setting using an off-the-shelf mask proposal model in~\secref{sec:off_the_shelf_masks}.

\textbf{Default Settings for Ablations}\quad
Unless otherwise specified, we use the default setting below for ablation studies: We resize both the image and mask during training until the longer side reaches a length of $896$ pixels, and then pad the shorter side to match this length.
We apply minimal data augmentation, limited to random flipping.
The context queries in \bridgename attend to the whole image.
We initialize \modelname with InstructBLIP~\cite{instructblip} pre-trained weight, which uses EVA-ViT-g/224~\cite{fang2023eva} as vision encoder, and Vicuna-7B~\cite{vicuna2023} as LLM. We use $32$ mask queries, $32$ context queries, and $1$ mode query which is randomly selected between vocab-agnostic query (shared across datasets) and vocab-specific query (one per dataset).

\textbf{Settings for Final Models}\quad
Based on the findings in the ablation studies (we detail the results later), for our final model, we increase the image resolution to $1120$ and the context queries attend to the bounding box region that is $0.5\times$ larger than the box-constrained mask region.
We also use random scale jittering in the range of $[0.5, 1.5]$.

\begin{table*}[t]
\centering
\small
\tablestyle{3pt}{1.2}
\begin{tabular}{cccccccccccccc|cc}
 \multicolumn{2}{c}{} & \multicolumn{2}{c}{COCO} & \multicolumn{2}{c}{LVIS} & \multicolumn{2}{c}{ADE20K} & \multicolumn{2}{c}{Cityscapes} & \multicolumn{2}{c}{A847} & \multicolumn{2}{c}{PC459} & \multicolumn{2}{c}{Avg} \\
 \multicolumn{2}{c}{methods} & Acc &  NIV & Acc &  NIV & Acc & NIV & Acc & NIV & Acc & NIV & Acc & NIV & Acc & NIV \\

\shline
\multicolumn{16}{c}{Single Dataset} \\
\shline

\multicolumn{2}{c}{COCO Only} & $85.5$ &  $5.3e^{-5}$ & - &  - & - & - & - & - & - & - & - & - & - & -\\
\multicolumn{2}{c}{LVIS Only} & - &  - & $68.3$ &  $2.4e^{-3}$ & - & - & - & - & - & - & - & - & - & -\\
\multicolumn{2}{c}{ADE20K Only} & - &  - & - &  - & $82.3$ & $4.3e^{-4}$ & - & - & - & - & - & - & - & -\\
\multicolumn{2}{c}{Cityscapes Only} & - &  - & - &  - & - & - & $79.4$ & $4.9e^{-4}$ & - & - & - & - & - & -\\
\multicolumn{2}{c}{A847 Only} & - &  - & - &  - & - & - & - & - & $76.9$ & $3.3e^{-3}$ & - & - & - & -\\
\multicolumn{2}{c}{PC459 Only} & - &  - & - &  - & - & - & - & - & - & - & $80.9$ & $6.9e^{-3}$ & - & -\\

\shline
\multicolumn{16}{c}{Multiple Datasets} \\
\shline
\multicolumn{2}{c}{Learnable Embed} & $84.3$ & $0.00$ & $67.3$ & $0.00$ & $82.3$ & $0.00$ & $81.0$ & $0.00$ & $76.0$ & $0.00$ & $82.5$ & $0.00$ & $78.9$  & $0.00$\\
\multicolumn{2}{c}{Text Embed}  & $83.4$ & $0.00$ & $65.4$ & $0.00$ & $81.5$ & $0.00$ & $81.6$ & $0.00$ & $75.1$ & $0.00$ & $81.7$ & $0.00$ & $78.1$ & $0.00$ \\

\multirow{ 2}{*}{\modelname} & vocab-agnostic & $74.9$ & $11.1$ & $56.8$ & $10.0$ & $80.6$ & $2.31$ & $81.1$ & $0.01$ & $75.6$ & $0.60$ & $77.8$ & $5.69$ & $74.5$ & $4.95$\\
& vocab-specific & $84.7$ & $0.10$ & $67.0$ & $0.62$ & $82.1$ & $0.55$ & $81.1$ & $7.0e^{-5}$ & $76.4$ & $0.49$ & $80.8$ & $1.90$ & $78.7$ & $0.61$\\
\multirow{ 2}{*}{\modelname$^\dagger$} & vocab-agnostic &$79.5$ & $8.75$ & $64.6$ & $8.22$ & $83.8$ & $2.11$ & $88.7$ & $0.01$ & $76.6$ & $0.81$ & $80.6$ & $3.74$ & $79.0$ & $3.94$\\
& vocab-specific & $87.0$ & $0.11$ & $72.7$ & $0.94$ & $85.2$ & $0.31$ & $88.6$ & $0.01$ & $78.1$ & $0.49$ & $83.0$ & $0.55$ & $82.4$ & $0.40$\\

\end{tabular}
\caption{
\textbf{Mask classification accuracy across the six segmentation datasets, using ground-truth masks.}
Note that \modelname (vocab-agnostic) and \modelname (vocab-specific) are obtained from the same model and weights, but activate vocabulary-agnostic or vocabulary-specific queries during inference, respectively.
NIV: Not-in-Vocabulary.
$\dagger$: Our final model setting.
}
\vspace{-1ex}
\label{tab:generative_ablation1}
\end{table*}

\subsection{Mask Classification with Ground-Truth Masks}
\label{sec:gt_masks}
\textbf{Generative Model for Discriminative Tasks}\quad
In~\tabref{tab:generative_ablation1}, we demonstrate that a generative model can effectively capture the training dataset, yielding predictions well-aligned with the training vocabulary.
Specifically, as shown in the top few rows of the table (``Single Dataset''), we first train \modelname separately on each of the six segmentation datasets, and evaluate its mask classification accuracy using ground-truth masks.
Remarkably, the model, although tasked with unrestricted generation of class names, consistently delivers predictions well within the vocabulary of its respective dataset.
This is evident by the high percentage of accurate predictions (\ie, high Acc scores) and the very low percentage of predictions falling outside the vocabulary (\ie, low NIV scores), showcasing the generative model's capacity to operate for a discriminative task.

Next, we explore the more interesting setting, where all six datasets are used for training (``Multiple Datasets'' in the table), where \modelname still maintains a high accuracy for each individual dataset, even in the presence of potential label conflicts.
Specifically, the proposed \queryname scheme effectively alleviates the label conflicts between datasets, where the vocabulary-specific queries (``vocab-specific'' in the table) better learns the associated vocabulary for each dataset, while the vocabulary-agnostic (``vocab-agnostic'') maintains the open-ended recognition ability (indicated by higher NIV scores).
Notably, this achievement is non-trivial and underscores the value of the proposed \queryname.

Additionally, we establish two discriminative baselines for comparisons.
The first one (denoted as ``Learnable Embed") replaces the frozen LLM with six learnable linear layers, each tailored to a specific dataset.
The second one (named ``Text Embed") initializes the classification layer with pre-extracted text embeddings and applies it individually to each dataset, approximating the approach presented in~\cite{gu2023dataseg, zhou2023lmseg}.
As shown in the table, our generative model \modelname performs comparably to the strong baseline ``Learnable Embed'' on the average (78.7\% \vs 78.9\% Acc) and outperforms the ``Text Embed'' baseline.
Our findings highlight that the generative model can perform on par with the discriminative models, even in discriminative tasks, underscoring its versatility and effectiveness.

Finally, in the last two rows of the table (denoted as \modelname~$\dagger$), using our final model setting (\eg, larger input size) can further significantly improve the performance for both vocabulary-agnostic and vocabulary-specific settings.

\begin{table*}[th]
\centering
\subfloat[
\textbf{Input Resolution}
\label{tab:input_res}
]{
\centering
\begin{minipage}{0.45\linewidth}{\begin{center}
\tablestyle{4pt}{1.1}
\begin{tabular}{c|ccccccc}
Input Res. & $224$ & $448$ & $672$ & $896$ & $1120$ & $1344$ & $1568$ \\
\shline
Avg Acc & $57.3$ & $70.1$ & $75.2$ & $78.7$ & $79.9$ & $76.6$ & $75.9$ \\
Avg NIV & $0.89$ & $0.84$ & $1.01$ & $0.61$ & $0.74$ 
 & $3.5$ & $3.5$\\
\end{tabular}
\end{center}}\end{minipage}
}
\subfloat[
\textbf{Sliding-Window Stride}
\label{tab:sliding_stride}
]{
\begin{minipage}{0.45\linewidth}{\begin{center}
\tablestyle{6pt}{1.1}
\begin{tabular}{c|cccc}
Sliding Stride & Global & $224$ & $168$ & $112$ \\
\shline
Avg Acc & $71.5$ & $78.7$ & $79.5$ & $79.6$ \\
Avg NIV & $0.68$ & $0.61$ & $0.59$ & $0.56$ \\
\end{tabular}
\end{center}}\end{minipage}
}
\\
\centering
\subfloat[
\textbf{\queryname}
\label{tab:vocab_specific}
]{
\begin{minipage}{0.45\linewidth}{\begin{center}
\tablestyle{4pt}{1.05}
\begin{tabular}{c|ccc}
vocab queries & None & Vocab-Agnostic & Vocab-Specific \\
\shline
Avg Acc & $74.8$ & $74.5$ & $78.7$ \\
Avg NIV & $4.45$ & $4.95$ & $0.61$ \\
\end{tabular}
\end{center}}\end{minipage}
}
\subfloat[
\textbf{Context Enlargement Ratio}
\label{tab:context_enlarge_ratio}
]{
\begin{minipage}{0.5\linewidth}{\begin{center}
\tablestyle{4pt}{1.05}
\begin{tabular}{c|cccccccc}
Ratio & Global & $0.0\times$ & $0.1\times$ & $0.2\times$ & $0.3\times$ & $0.4\times$ & $0.5\times$ & $0.6\times$ \\
\shline
Avg Acc & $78.7$ & $79.5$ & $80.5$ & $80.6$ & $80.9$ & $80.9$ & $81.3$ & $81.0$ \\
Avg NIV & $0.61$ & $0.51$ & $0.43$ & $0.50$ & $0.62$ & $0.56$ & $0.47$ & $0.50$ \\

\end{tabular}
\end{center}}\end{minipage}
}
\caption{
\textbf{Ablation studies on \modelname design choices.}
The ablated design choices include (a) image input resolution, (b) the sliding-window stride of the CLIP-ViT backbone to extract high-resolution image features, (c) employment of \queryname, and (d) the box region size attended by the context queries in \bridgename.
}
\label{tab:ablations} \vspace{-.5em}
\end{table*}

\textbf{Adaptation to Higher Input Resolution}\quad
In contrast to many multi-modal Large Language Models (LLM) approaches that directly employ the frozen CLIP-ViT, we emphasize the critical importance of higher input resolution for achieving accurate object-level recognition. However, we recognize that frozen Vision Transformers (ViTs) often exhibit inferior performance when adapting to larger input resolutions compared to their pre-training resolutions, as documented in~\cite{yu2023convolutions}. To address this limitation, we introduce a simple yet highly effective sliding-window approach for obtaining enhanced features from a frozen ViT when processing higher-resolution inputs.

As illustrated in~\tabref{tab:input_res}, our experiments consistently demonstrate performance gains as input resolution increases, particularly from $224\times224$ to $448\times448$, reflecting an impressive improvement of ($+12.8\%$). This underscores the pivotal role of a larger input resolution in achieving superior object-level recognition performance. The benefits persist until the input resolution reaches $1120\times1120$, while larger input resolution leads to a performance drop, potentially because each sliding-window fails to capture semantic meaningful feature. Notably, the ``Avg NIV" metric remains relatively stable across all experiments, indicating that the performance boost primarily stems from improved mask classification rather than a better overfitting with the respective vocabulary.

\textbf{Sliding-Window Stride}\quad
We validate our sliding-window design in~\tabref{tab:sliding_stride}, where direct application of the frozen ViT with high-resolution inputs (``Global") results in significantly inferior performance ($-7.2\%$), consistent with the observations in~\cite{yu2023convolutions}. Moreover, our findings reveal that employing the sliding-window approach with overlapping windows further enhances results, although the incremental benefit diminishes as the overlap increases. Considering the significant additional computational costs coming from overlapping window, we do not use it in our final setting.

\textbf{Effect of \queryname }\quad
It is evident from our experiments that the inclusion of \queryname plays a pivotal role in the effectiveness of \modelname. As demonstrated in~\tabref{tab:vocab_specific}, training \modelname across multiple datasets without these queries may result in better generalization capabilities but compromised alignment to specific datasets. This is evident through a lower ``Avg Acc" and a higher ``Avg NIV".
However, with the integration of the proposed \queryname, \modelname exhibits the ability to operate in both ``closed-ended" mode (vocab-specific) and ``open-ended" mode (vocab-agnostic). This allows the model to strike a balance between generalization and alignment, preserving both essential capabilities.

\textbf{Context is Important for Recognition}\quad
We investigate the critical role of context, as detailed in~\tabref{tab:context_enlarge_ratio}. Here, ``Global" signifies that the context attention may encompass the entire image, whereas ``$0.0\times$" refers to a tightly constrained bounding box that encircles the segmentation mask closely. The notation ``$k\times$" indicates the expansion of the bounding box by a factor of ``$k\times$" on each side.
The results in the table underscore the significance of context. Even a tightly defined bounding box offers a noteworthy improvement over global context ($+0.8\%$). Notably, the benefits become more pronounced as we progressively transition to a looser bounding box, with the most substantial gain occurring at ``$0.5\times$" ($+2.6\%$) compared to the global context configuration. This underlines the importance of context for accurate recognition, with an optimal balance between tight and loose bounding boxes yielding superior results.

\subsection{Mask Classification with Off-the-shelf Mask Proposal Model}
\label{sec:off_the_shelf_masks}

\textbf{Benchmarking with Other Generalists}\quad
In addition to evaluating \modelname with ground-truth masks, we also provide a practical assessment by integrating \modelname with an off-the-shelf mask proposal model. 
We employ mask proposals generated by kMaX-DeepLab~\cite{yu2022k} and then apply \modelname for classifying these mask proposals. 
We focus on the comparisons with other generalist segmentation models that are jointly trained with multiple segmentation datasets, similar to our setting.
Specifically, we compare with text embedding-based methods like LMSeg~\cite{zhou2023lmseg} and DaTaSeg~\cite{gu2023dataseg} across various datasets, including COCO panoptic, ADE20K panoptic and semantic, Cityscapes panoptic, and semantic segmentation.
As outlined in~\tabref{tab:final_comparison}, \modelname consistently achieves higher Panoptic Quality (PQ) or mean Intersection-over-Union (mIoU) scores in comparison to other discriminative methods.  Specifically, with ResNet50 proposal model backbone, \modelname outperforms LMSeg~\cite{zhou2023lmseg} by $+14.7$, $+8.4$, and $+4.7$ PQ on COCO, ADE20K, Cityscapes respectively. Compared to DaTaSeg~\cite{gu2023dataseg}, \modelname also improve the COCO PQ by $+4.3$, $+2.6$, the ADE20K mIoU by $+1.9$, $+1.2$ for ResNet50 and Large backbone variants, respectively. \modelname also shows comparable performance to the specialist model Mask2Former~\cite{cheng2022masked}.

\begin{table}[t]
\centering
\tablestyle{2pt}{1.2}
\begin{tabular}{lc|c|cc|cc}
 & proposal network & \multicolumn{1}{c|}{COCO} & \multicolumn{2}{c|}{ADE20K} & \multicolumn{2}{c}{Cityscapes} \\
methods &  backbone & PQ & PQ & mIoU & PQ & mIoU \\

\shline
\multicolumn{7}{c}{Specialist Models (one model per dataset)} \\
\shline
Mask2Former~\cite{cheng2022masked} & ResNet50~\cite{he2016deep} & 51.9 & 39.7 & 46.1 & 62.1 & 77.5 \\
Mask2Former~\cite{cheng2022masked} & Swin-L~\cite{liu2021swin} & 57.8 & 48.1 & 54.5 & 66.6 & 82.9 \\
\shline
\multicolumn{7}{c}{Generalist Models (one model for all datasets)} \\
\shline
LMSeg~\cite{zhou2023lmseg} & ResNet50~\cite{he2016deep} & 38.6 & 35.4 & 45.2 & 54.8 & 80.9 \\
DaTaSeg~\cite{gu2023dataseg} & ResNet50~\cite{he2016deep} & 49.0 &
\textcolor{datasegcolor}{29.8} & 48.1 & - & - \\
DaTaSeg~\cite{gu2023dataseg} & ViTDet-L~\cite{li2022exploring} & 53.5 & \textcolor{datasegcolor}{33.4} & 54.0 & - & - \\
\shline
\modelname & ResNet50~\cite{he2016deep} & 53.3 & 43.8 & 50.0 & 59.5 & 77.0  \\
\modelname & ConvNeXt-L~\cite{liu2022convnet} & 56.1 & 49.7 & 55.2 & 64.7 & 80.2  \\
\end{tabular}
\caption{
\textbf{Comparison of \modelname and other discrminative models.}
\modelname uses kMaX-DeepLab~\cite{yu2022k} as the mask proposal model.
We mainly compare with the generalist models (LMSeg and DaTaSeg) and list Mask2Former as the specialist model for reference.
Note that DaTaSeg~\cite{gu2023dataseg} uses ADE20K-semantic training data instead of ADE20K-panoptic (thus mark in gray).}
\vspace{-3ex}
\label{tab:final_comparison}
\end{table}
\vspace{-1ex}
\section{Conclusions}
\label{sec:conclusions}
In this study, we introduced a novel challenge in the domain of visual recognition, referred to as open-ended visual recognition, and introduced \modelname, a generative framework designed to address this challenge. \modelname is a mask-aware multi-modal LLM capable of processing segmentation masks as inputs and generating semantic class predictions in a generative manner, without relying on a predefined vocabulary. Our empirical findings demonstrate that this generative model yields promising recognition accuracy and exhibits significant potential for real-world applications, particularly in handling novel concepts that extend beyond predefined vocabularies.
{
    \small
    \bibliographystyle{ieeenat_fullname}
    \bibliography{main}

\begin{thebibliography}{92}
\providecommand{\natexlab}[1]{#1}
\providecommand{\url}[1]{\texttt{#1}}
\expandafter\ifx\csname urlstyle\endcsname\relax
  \providecommand{\doi}[1]{doi: #1}\else
  \providecommand{\doi}{doi: \begingroup \urlstyle{rm}\Url}\fi

\bibitem[Alayrac et~al.(2022)Alayrac, Donahue, Luc, Miech, Barr, Hasson, Lenc, Mensch, Millican, Reynolds, et~al.]{alayrac2022flamingo}
Jean-Baptiste Alayrac, Jeff Donahue, Pauline Luc, Antoine Miech, Iain Barr, Yana Hasson, Karel Lenc, Arthur Mensch, Katherine Millican, Malcolm Reynolds, et~al.
\newblock Flamingo: a visual language model for few-shot learning.
\newblock \emph{NeurIPS}, 2022.

\bibitem[Brown et~al.(2020)Brown, Mann, Ryder, Subbiah, Kaplan, Dhariwal, Neelakantan, Shyam, Sastry, Askell, et~al.]{brown2020language}
Tom Brown, Benjamin Mann, Nick Ryder, Melanie Subbiah, Jared~D Kaplan, Prafulla Dhariwal, Arvind Neelakantan, Pranav Shyam, Girish Sastry, Amanda Askell, et~al.
\newblock Language models are few-shot learners.
\newblock \emph{NeurIPS}, 2020.

\bibitem[Carion et~al.(2020)Carion, Massa, Synnaeve, Usunier, Kirillov, and Zagoruyko]{carion2020end}
Nicolas Carion, Francisco Massa, Gabriel Synnaeve, Nicolas Usunier, Alexander Kirillov, and Sergey Zagoruyko.
\newblock End-to-end object detection with transformers.
\newblock In \emph{ECCV}, 2020.

\bibitem[Chen et~al.(2023{\natexlab{a}})Chen, Zhu, Shen, Li, Liu, Zhang, Krishnamoorthi, Chandra, Xiong, and Elhoseiny]{chen2023minigptv2}
Jun Chen, Deyao Zhu, Xiaoqian Shen, Xiang Li, Zechu Liu, Pengchuan Zhang, Raghuraman Krishnamoorthi, Vikas Chandra, Yunyang Xiong, and Mohamed Elhoseiny.
\newblock Minigpt-v2: large language model as a unified interface for vision-language multi-task learning.
\newblock \emph{arXiv preprint arXiv:2310.09478}, 2023{\natexlab{a}}.

\bibitem[Chen et~al.(2023{\natexlab{b}})Chen, Zhang, Zeng, Zhang, Zhu, and Zhao]{chen2023shikra}
Keqin Chen, Zhao Zhang, Weili Zeng, Richong Zhang, Feng Zhu, and Rui Zhao.
\newblock Shikra: Unleashing multimodal llm's referential dialogue magic.
\newblock \emph{arXiv preprint arXiv:2306.15195}, 2023{\natexlab{b}}.

\bibitem[Chen et~al.(2015{\natexlab{a}})Chen, Papandreou, Kokkinos, Murphy, and Yuille]{chen015deeplabv1}
Liang-Chieh Chen, George Papandreou, Iasonas Kokkinos, Kevin Murphy, and Alan Yuille.
\newblock Semantic image segmentation with deep convolutional nets and fully connected crfs.
\newblock In \emph{ICLR}, 2015{\natexlab{a}}.

\bibitem[Chen et~al.(2017{\natexlab{a}})Chen, Papandreou, Kokkinos, Murphy, and Yuille]{chen2017deeplab}
Liang-Chieh Chen, George Papandreou, Iasonas Kokkinos, Kevin Murphy, and Alan~L Yuille.
\newblock Deeplab: Semantic image segmentation with deep convolutional nets, atrous convolution, and fully connected crfs.
\newblock \emph{TPAMI}, 2017{\natexlab{a}}.

\bibitem[Chen et~al.(2017{\natexlab{b}})Chen, Papandreou, Schroff, and Adam]{chen2017rethinking}
Liang-Chieh Chen, George Papandreou, Florian Schroff, and Hartwig Adam.
\newblock Rethinking atrous convolution for semantic image segmentation.
\newblock \emph{arXiv preprint arXiv:1706.05587}, 2017{\natexlab{b}}.

\bibitem[Chen et~al.(2018)Chen, Zhu, Papandreou, Schroff, and Adam]{chen2018encoder}
Liang-Chieh Chen, Yukun Zhu, George Papandreou, Florian Schroff, and Hartwig Adam.
\newblock Encoder-decoder with atrous separable convolution for semantic image segmentation.
\newblock In \emph{ECCV}, 2018.

\bibitem[Chen et~al.(2014)Chen, Mottaghi, Liu, Fidler, Urtasun, and Yuille]{chen2014detect}
Xianjie Chen, Roozbeh Mottaghi, Xiaobai Liu, Sanja Fidler, Raquel Urtasun, and Alan Yuille.
\newblock Detect what you can: Detecting and representing objects using holistic models and body parts.
\newblock In \emph{CVPR}, 2014.

\bibitem[Chen et~al.(2015{\natexlab{b}})Chen, Fang, Lin, Vedantam, Gupta, Doll{\'a}r, and Zitnick]{chen2015microsoft}
Xinlei Chen, Hao Fang, Tsung-Yi Lin, Ramakrishna Vedantam, Saurabh Gupta, Piotr Doll{\'a}r, and C~Lawrence Zitnick.
\newblock Microsoft coco captions: Data collection and evaluation server.
\newblock \emph{arXiv preprint arXiv:1504.00325}, 2015{\natexlab{b}}.

\bibitem[Cheng et~al.(2021)Cheng, Schwing, and Kirillov]{cheng2021per}
Bowen Cheng, Alex Schwing, and Alexander Kirillov.
\newblock Per-pixel classification is not all you need for semantic segmentation.
\newblock \emph{NeurIPS}, 2021.

\bibitem[Cheng et~al.(2022)Cheng, Misra, Schwing, Kirillov, and Girdhar]{cheng2022masked}
Bowen Cheng, Ishan Misra, Alexander~G Schwing, Alexander Kirillov, and Rohit Girdhar.
\newblock Masked-attention mask transformer for universal image segmentation.
\newblock In \emph{CVPR}, 2022.

\bibitem[Chiang et~al.(2023)Chiang, Li, Lin, Sheng, Wu, Zhang, Zheng, Zhuang, Zhuang, Gonzalez, Stoica, and Xing]{vicuna2023}
Wei-Lin Chiang, Zhuohan Li, Zi Lin, Ying Sheng, Zhanghao Wu, Hao Zhang, Lianmin Zheng, Siyuan Zhuang, Yonghao Zhuang, Joseph~E. Gonzalez, Ion Stoica, and Eric~P. Xing.
\newblock Vicuna: An open-source chatbot impressing gpt-4 with 90\%* chatgpt quality, 2023.

\bibitem[Chung et~al.(2022)Chung, Hou, Longpre, Zoph, Tay, Fedus, Li, Wang, Dehghani, Brahma, et~al.]{chung2022scaling}
Hyung~Won Chung, Le Hou, Shayne Longpre, Barret Zoph, Yi Tay, William Fedus, Eric Li, Xuezhi Wang, Mostafa Dehghani, Siddhartha Brahma, et~al.
\newblock Scaling instruction-finetuned language models.
\newblock \emph{arXiv preprint arXiv:2210.11416}, 2022.

\bibitem[Conti et~al.(2023)Conti, Fini, Mancini, Rota, Wang, and Ricci]{conti2023vocabulary}
Alessandro Conti, Enrico Fini, Massimiliano Mancini, Paolo Rota, Yiming Wang, and Elisa Ricci.
\newblock Vocabulary-free image classification.
\newblock \emph{NeurIPS}, 2023.

\bibitem[Cordts et~al.(2016)Cordts, Omran, Ramos, Rehfeld, Enzweiler, Benenson, Franke, Roth, and Schiele]{cordts2016cityscapes}
Marius Cordts, Mohamed Omran, Sebastian Ramos, Timo Rehfeld, Markus Enzweiler, Rodrigo Benenson, Uwe Franke, Stefan Roth, and Bernt Schiele.
\newblock The cityscapes dataset for semantic urban scene understanding.
\newblock In \emph{CVPR}, 2016.

\bibitem[Dai et~al.(2023)Dai, Li, Li, Tiong, Zhao, Wang, Li, Fung, and Hoi]{instructblip}
Wenliang Dai, Junnan Li, Dongxu Li, Anthony Meng~Huat Tiong, Junqi Zhao, Weisheng Wang, Boyang Li, Pascale Fung, and Steven Hoi.
\newblock Instructblip: Towards general-purpose vision-language models with instruction tuning.
\newblock \emph{arXiv preprint arXiv:2305.06500}, 2023.

\bibitem[Deng et~al.(2009)Deng, Dong, Socher, Li, Li, and Fei-Fei]{deng2009imagenet}
Jia Deng, Wei Dong, Richard Socher, Li-Jia Li, Kai Li, and Li Fei-Fei.
\newblock Imagenet: A large-scale hierarchical image database.
\newblock In \emph{CVPR}, 2009.

\bibitem[Devlin et~al.(2018)Devlin, Chang, Lee, and Toutanova]{devlin2018bert}
Jacob Devlin, Ming-Wei Chang, Kenton Lee, and Kristina Toutanova.
\newblock Bert: Pre-training of deep bidirectional transformers for language understanding.
\newblock \emph{NAACL}, 2018.

\bibitem[Ding et~al.(2023)Ding, Wang, and Tu]{ding2022open}
Zheng Ding, Jieke Wang, and Zhuowen Tu.
\newblock Open-vocabulary panoptic segmentation with maskclip.
\newblock In \emph{ICML}, 2023.

\bibitem[Dosovitskiy et~al.(2021)Dosovitskiy, Beyer, Kolesnikov, Weissenborn, Zhai, Unterthiner, Dehghani, Minderer, Heigold, Gelly, Uszkoreit, and Houlsby]{dosovitskiy2020image}
Alexey Dosovitskiy, Lucas Beyer, Alexander Kolesnikov, Dirk Weissenborn, Xiaohua Zhai, Thomas Unterthiner, Mostafa Dehghani, Matthias Minderer, Georg Heigold, Sylvain Gelly, Jakob Uszkoreit, and Neil Houlsby.
\newblock An image is worth 16x16 words: Transformers for image recognition at scale.
\newblock In \emph{ICLR}, 2021.

\bibitem[Everingham et~al.(2010)Everingham, Van~Gool, Williams, Winn, and Zisserman]{everingham2010pascal}
Mark Everingham, Luc Van~Gool, Christopher~KI Williams, John Winn, and Andrew Zisserman.
\newblock The pascal visual object classes (voc) challenge.
\newblock \emph{IJCV}, 2010.

\bibitem[Fang et~al.(2023)Fang, Wang, Xie, Sun, Wu, Wang, Huang, Wang, and Cao]{fang2023eva}
Yuxin Fang, Wen Wang, Binhui Xie, Quan Sun, Ledell Wu, Xinggang Wang, Tiejun Huang, Xinlong Wang, and Yue Cao.
\newblock Eva: Exploring the limits of masked visual representation learning at scale.
\newblock In \emph{CVPR}, 2023.

\bibitem[Ghiasi et~al.(2022)Ghiasi, Gu, Cui, and Lin]{ghiasi2022scaling}
Golnaz Ghiasi, Xiuye Gu, Yin Cui, and Tsung-Yi Lin.
\newblock Scaling open-vocabulary image segmentation with image-level labels.
\newblock In \emph{ECCV}, 2022.

\bibitem[Girshick(2015)]{girshick2015fast}
Ross Girshick.
\newblock Fast r-cnn.
\newblock In \emph{ICCV}, 2015.

\bibitem[Goyal et~al.(2017)Goyal, Khot, Summers-Stay, Batra, and Parikh]{goyal2017making}
Yash Goyal, Tejas Khot, Douglas Summers-Stay, Dhruv Batra, and Devi Parikh.
\newblock Making the v in vqa matter: Elevating the role of image understanding in visual question answering.
\newblock In \emph{CVPR}, 2017.

\bibitem[Gu et~al.(2023)Gu, Cui, Huang, Rashwan, Yang, Zhou, Ghiasi, Kuo, Chen, Chen, and Ross]{gu2023dataseg}
Xiuye Gu, Yin Cui, Jonathan Huang, Abdullah Rashwan, Xuan Yang, Xingyi Zhou, Golnaz Ghiasi, Weicheng Kuo, Huizhong Chen, Liang-Chieh Chen, and David Ross.
\newblock Dataseg: Taming a universal multi-dataset multi-task segmentation model.
\newblock \emph{NeurIPS}, 2023.

\bibitem[Gupta et~al.(2019)Gupta, Dollar, and Girshick]{gupta2019lvis}
Agrim Gupta, Piotr Dollar, and Ross Girshick.
\newblock Lvis: A dataset for large vocabulary instance segmentation.
\newblock In \emph{CVPR}, 2019.

\bibitem[He et~al.(2022)He, Yang, Yang, Kortylewski, Yuan, Chen, Liu, Yang, Yu, and Yuille]{he2022partimagenet}
Ju He, Shuo Yang, Shaokang Yang, Adam Kortylewski, Xiaoding Yuan, Jie-Neng Chen, Shuai Liu, Cheng Yang, Qihang Yu, and Alan Yuille.
\newblock Partimagenet: A large, high-quality dataset of parts.
\newblock In \emph{ECCV}, 2022.

\bibitem[He et~al.(2016)He, Zhang, Ren, and Sun]{he2016deep}
Kaiming He, Xiangyu Zhang, Shaoqing Ren, and Jian Sun.
\newblock Deep residual learning for image recognition.
\newblock In \emph{CVPR}, 2016.

\bibitem[He et~al.(2017)He, Gkioxari, Doll{\'a}r, and Girshick]{he2017mask}
Kaiming He, Georgia Gkioxari, Piotr Doll{\'a}r, and Ross Girshick.
\newblock Mask r-cnn.
\newblock In \emph{ICCV}, 2017.

\bibitem[Hendrycks et~al.(2021)Hendrycks, Burns, Basart, Zou, Mazeika, Song, and Steinhardt]{hendrycks2020measuring}
Dan Hendrycks, Collin Burns, Steven Basart, Andy Zou, Mantas Mazeika, Dawn Song, and Jacob Steinhardt.
\newblock Measuring massive multitask language understanding.
\newblock In \emph{ICLR}, 2021.

\bibitem[Hudson and Manning(2019)]{hudson2019gqa}
Drew~A Hudson and Christopher~D Manning.
\newblock Gqa: A new dataset for real-world visual reasoning and compositional question answering.
\newblock In \emph{CVPR}, 2019.

\bibitem[Jia et~al.(2021)Jia, Yang, Xia, Chen, Parekh, Pham, Le, Sung, Li, and Duerig]{jia2021scaling}
Chao Jia, Yinfei Yang, Ye Xia, Yi-Ting Chen, Zarana Parekh, Hieu Pham, Quoc Le, Yun-Hsuan Sung, Zhen Li, and Tom Duerig.
\newblock Scaling up visual and vision-language representation learning with noisy text supervision.
\newblock In \emph{ICML}, 2021.

\bibitem[Kim et~al.(2022)Kim, Lin, Angelova, Kweon, and Kuo]{kim2022learning}
Dahun Kim, Tsung-Yi Lin, Anelia Angelova, In~So Kweon, and Weicheng Kuo.
\newblock Learning open-world object proposals without learning to classify.
\newblock In \emph{ICRA}, 2022.

\bibitem[Kingma and Ba(2015)]{kingma2014adam}
Diederik~P Kingma and Jimmy Ba.
\newblock Adam: A method for stochastic optimization.
\newblock In \emph{ICLR}, 2015.

\bibitem[Kirillov et~al.(2019)Kirillov, He, Girshick, Rother, and Doll{\'a}r]{kirillov2019panoptic}
Alexander Kirillov, Kaiming He, Ross Girshick, Carsten Rother, and Piotr Doll{\'a}r.
\newblock Panoptic segmentation.
\newblock In \emph{CVPR}, 2019.

\bibitem[Kirillov et~al.(2023)Kirillov, Mintun, Ravi, Mao, Rolland, Gustafson, Xiao, Whitehead, Berg, Lo, Doll{\'a}r, and Girshick]{kirillov2023sa-1b}
Alexander Kirillov, Eric Mintun, Nikhila Ravi, Hanzi Mao, Chloe Rolland, Laura Gustafson, Tete Xiao, Spencer Whitehead, Alexander~C. Berg, Wan-Yen Lo, Piotr Doll{\'a}r, and Ross Girshick.
\newblock Segment anything.
\newblock In \emph{ICCV}, 2023.

\bibitem[Krizhevsky et~al.(2012)Krizhevsky, Sutskever, and Hinton]{krizhevsky2012imagenet}
Alex Krizhevsky, Ilya Sutskever, and Geoffrey~E Hinton.
\newblock Imagenet classification with deep convolutional neural networks.
\newblock \emph{NeurIPS}, 2012.

\bibitem[Lambert et~al.(2020)Lambert, Liu, Sener, Hays, and Koltun]{lambert2020mseg}
John Lambert, Zhuang Liu, Ozan Sener, James Hays, and Vladlen Koltun.
\newblock Mseg: A composite dataset for multi-domain semantic segmentation.
\newblock In \emph{CVPR}, 2020.

\bibitem[Li et~al.(2022{\natexlab{a}})Li, Li, Xiong, and Hoi]{li2022blip}
Junnan Li, Dongxu Li, Caiming Xiong, and Steven Hoi.
\newblock Blip: Bootstrapping language-image pre-training for unified vision-language understanding and generation.
\newblock In \emph{ICML}, 2022{\natexlab{a}}.

\bibitem[Li et~al.(2023)Li, Li, Savarese, and Hoi]{li2023blip}
Junnan Li, Dongxu Li, Silvio Savarese, and Steven Hoi.
\newblock Blip-2: Bootstrapping language-image pre-training with frozen image encoders and large language models.
\newblock In \emph{ICML}, 2023.

\bibitem[Li and Liang(2021)]{li2021prefix}
Xiang~Lisa Li and Percy Liang.
\newblock Prefix-tuning: Optimizing continuous prompts for generation.
\newblock In \emph{ACL}, 2021.

\bibitem[Li et~al.(2022{\natexlab{b}})Li, Mao, Girshick, and He]{li2022exploring}
Yanghao Li, Hanzi Mao, Ross Girshick, and Kaiming He.
\newblock Exploring plain vision transformer backbones for object detection.
\newblock In \emph{ECCV}, 2022{\natexlab{b}}.

\bibitem[Liang et~al.(2023)Liang, Wu, Dai, Li, Zhao, Zhang, Zhang, Vajda, and Marculescu]{liang2023open}
Feng Liang, Bichen Wu, Xiaoliang Dai, Kunpeng Li, Yinan Zhao, Hang Zhang, Peizhao Zhang, Peter Vajda, and Diana Marculescu.
\newblock Open-vocabulary semantic segmentation with mask-adapted clip.
\newblock In \emph{CVPR}, 2023.

\bibitem[Lin et~al.(2014)Lin, Maire, Belongie, Hays, Perona, Ramanan, Doll{\'a}r, and Zitnick]{lin2014microsoft}
Tsung-Yi Lin, Michael Maire, Serge Belongie, James Hays, Pietro Perona, Deva Ramanan, Piotr Doll{\'a}r, and C~Lawrence Zitnick.
\newblock Microsoft coco: Common objects in context.
\newblock In \emph{ECCV}, 2014.

\bibitem[Liu et~al.(2023{\natexlab{a}})Liu, Li, Li, and Lee]{liu2023improved}
Haotian Liu, Chunyuan Li, Yuheng Li, and Yong~Jae Lee.
\newblock Improved baselines with visual instruction tuning.
\newblock \emph{arXiv preprint arXiv:2310.03744}, 2023{\natexlab{a}}.

\bibitem[Liu et~al.(2023{\natexlab{b}})Liu, Li, Wu, and Lee]{liu2023visual}
Haotian Liu, Chunyuan Li, Qingyang Wu, and Yong~Jae Lee.
\newblock Visual instruction tuning.
\newblock \emph{NeurIPS}, 2023{\natexlab{b}}.

\bibitem[Liu et~al.(2021)Liu, Lin, Cao, Hu, Wei, Zhang, Lin, and Guo]{liu2021swin}
Ze Liu, Yutong Lin, Yue Cao, Han Hu, Yixuan Wei, Zheng Zhang, Stephen Lin, and Baining Guo.
\newblock Swin transformer: Hierarchical vision transformer using shifted windows.
\newblock In \emph{ICCV}, 2021.

\bibitem[Liu et~al.(2022)Liu, Mao, Wu, Feichtenhofer, Darrell, and Xie]{liu2022convnet}
Zhuang Liu, Hanzi Mao, Chao-Yuan Wu, Christoph Feichtenhofer, Trevor Darrell, and Saining Xie.
\newblock A convnet for the 2020s.
\newblock In \emph{CVPR}, 2022.

\bibitem[Long et~al.(2015)Long, Shelhamer, and Darrell]{long2015fully}
Jonathan Long, Evan Shelhamer, and Trevor Darrell.
\newblock Fully convolutional networks for semantic segmentation.
\newblock In \emph{CVPR}, 2015.

\bibitem[Loshchilov and Hutter(2017)]{loshchilov2017decoupled}
Ilya Loshchilov and Frank Hutter.
\newblock Decoupled weight decay regularization.
\newblock \emph{arXiv preprint arXiv:1711.05101}, 2017.

\bibitem[Nguyen et~al.(2023)Nguyen, Gadre, Ilharco, Oh, and Schmidt]{nguyen2023improving}
Thao Nguyen, Samir~Yitzhak Gadre, Gabriel Ilharco, Sewoong Oh, and Ludwig Schmidt.
\newblock Improving multimodal datasets with image captioning.
\newblock \emph{arXiv preprint arXiv:2307.10350}, 2023.

\bibitem[OpenAI(2023)]{openai2023gpt}
OpenAI.
\newblock Gpt-4 technical report.
\newblock \emph{arXiv preprint arXiv:2303.08774}, 2023.

\bibitem[Ouyang-Zhang et~al.(2022)Ouyang-Zhang, Cho, Zhou, and Kr{\"a}henb{\"u}hl]{ouyang2022nms}
Jeffrey Ouyang-Zhang, Jang~Hyun Cho, Xingyi Zhou, and Philipp Kr{\"a}henb{\"u}hl.
\newblock Nms strikes back.
\newblock \emph{arXiv preprint arXiv:2212.06137}, 2022.

\bibitem[Peng et~al.(2023)Peng, Wang, Dong, Hao, Huang, Ma, and Wei]{peng2023kosmos}
Zhiliang Peng, Wenhui Wang, Li Dong, Yaru Hao, Shaohan Huang, Shuming Ma, and Furu Wei.
\newblock Kosmos-2: Grounding multimodal large language models to the world.
\newblock \emph{arXiv preprint arXiv:2306.14824}, 2023.

\bibitem[Plummer et~al.(2015)Plummer, Wang, Cervantes, Caicedo, Hockenmaier, and Lazebnik]{plummer2015flickr30k}
Bryan~A Plummer, Liwei Wang, Chris~M Cervantes, Juan~C Caicedo, Julia Hockenmaier, and Svetlana Lazebnik.
\newblock Flickr30k entities: Collecting region-to-phrase correspondences for richer image-to-sentence models.
\newblock In \emph{ICCV}, 2015.

\bibitem[Qin et~al.(2023)Qin, Wu, Yan, Li, Yuxi, Xiao, Wang, Wang, Wen, Pan, and Wang]{qin2023freeseg}
Jie Qin, Jie Wu, Pengxiang Yan, Ming Li, Ren Yuxi, Xuefeng Xiao, Yitong Wang, Rui Wang, Shilei Wen, Xin Pan, and Xingang Wang.
\newblock Freeseg: Unified, universal and open-vocabulary image segmentation.
\newblock In \emph{CVPR}, 2023.

\bibitem[Radford et~al.(2019)Radford, Wu, Child, Luan, Amodei, and Sutskever]{radford2019language}
Alec Radford, Jeffrey Wu, Rewon Child, David Luan, Dario Amodei, and Ilya Sutskever.
\newblock Language models are unsupervised multitask learners.
\newblock \emph{OpenAI blog}, 2019.

\bibitem[Radford et~al.(2021)Radford, Kim, Hallacy, Ramesh, Goh, Agarwal, Sastry, Askell, Mishkin, Clark, Krueger, and Sutskever]{radford2021learning}
Alec Radford, Jong~Wook Kim, Chris Hallacy, Aditya Ramesh, Gabriel Goh, Sandhini Agarwal, Girish Sastry, Amanda Askell, Pamela Mishkin, Jack Clark, Gretchen Krueger, and Ilya Sutskever.
\newblock Learning transferable visual models from natural language supervision.
\newblock In \emph{ICML}, 2021.

\bibitem[Raffel et~al.(2020)Raffel, Shazeer, Roberts, Lee, Narang, Matena, Zhou, Li, and Liu]{raffel2020exploring}
Colin Raffel, Noam Shazeer, Adam Roberts, Katherine Lee, Sharan Narang, Michael Matena, Yanqi Zhou, Wei Li, and Peter~J Liu.
\newblock Exploring the limits of transfer learning with a unified text-to-text transformer.
\newblock \emph{The Journal of Machine Learning Research}, 2020.

\bibitem[Roberts et~al.(2020)Roberts, Raffel, and Shazeer]{roberts2020much}
Adam Roberts, Colin Raffel, and Noam Shazeer.
\newblock How much knowledge can you pack into the parameters of a language model?
\newblock In \emph{EMNLP}, 2020.

\bibitem[Schuhmann et~al.(2022)Schuhmann, Beaumont, Vencu, Gordon, Wightman, Cherti, Coombes, Katta, Mullis, Wortsman, et~al.]{schuhmann2022laion}
Christoph Schuhmann, Romain Beaumont, Richard Vencu, Cade Gordon, Ross Wightman, Mehdi Cherti, Theo Coombes, Aarush Katta, Clayton Mullis, Mitchell Wortsman, et~al.
\newblock Laion-5b: An open large-scale dataset for training next generation image-text models.
\newblock \emph{NeurIPS}, 2022.

\bibitem[Sutskever et~al.(2014)Sutskever, Vinyals, and Le]{sutskever2014sequence}
Ilya Sutskever, Oriol Vinyals, and Quoc~V Le.
\newblock Sequence to sequence learning with neural networks.
\newblock \emph{NeurIPS}, 2014.

\bibitem[Thoppilan et~al.(2022)Thoppilan, De~Freitas, Hall, Shazeer, Kulshreshtha, Cheng, Jin, Bos, Baker, Du, et~al.]{thoppilan2022lamda}
Romal Thoppilan, Daniel De~Freitas, Jamie Hall, Noam Shazeer, Apoorv Kulshreshtha, Heng-Tze Cheng, Alicia Jin, Taylor Bos, Leslie Baker, Yu Du, et~al.
\newblock Lamda: Language models for dialog applications.
\newblock \emph{arXiv preprint arXiv:2201.08239}, 2022.

\bibitem[Touvron et~al.(2023{\natexlab{a}})Touvron, Lavril, Izacard, Martinet, Lachaux, Lacroix, Rozi{\`e}re, Goyal, Hambro, Azhar, et~al.]{touvron2023llama}
Hugo Touvron, Thibaut Lavril, Gautier Izacard, Xavier Martinet, Marie-Anne Lachaux, Timoth{\'e}e Lacroix, Baptiste Rozi{\`e}re, Naman Goyal, Eric Hambro, Faisal Azhar, et~al.
\newblock Llama: Open and efficient foundation language models.
\newblock \emph{arXiv preprint arXiv:2302.13971}, 2023{\natexlab{a}}.

\bibitem[Touvron et~al.(2023{\natexlab{b}})Touvron, Martin, Stone, Albert, Almahairi, Babaei, Bashlykov, Batra, Bhargava, Bhosale, et~al.]{touvron2023llama2}
Hugo Touvron, Louis Martin, Kevin Stone, Peter Albert, Amjad Almahairi, Yasmine Babaei, Nikolay Bashlykov, Soumya Batra, Prajjwal Bhargava, Shruti Bhosale, et~al.
\newblock Llama 2: Open foundation and fine-tuned chat models.
\newblock \emph{arXiv preprint arXiv:2307.09288}, 2023{\natexlab{b}}.

\bibitem[Vaswani et~al.(2017)Vaswani, Shazeer, Parmar, Uszkoreit, Jones, Gomez, Kaiser, and Polosukhin]{vaswani2017attention}
Ashish Vaswani, Noam Shazeer, Niki Parmar, Jakob Uszkoreit, Llion Jones, Aidan~N Gomez, {\L}ukasz Kaiser, and Illia Polosukhin.
\newblock Attention is all you need.
\newblock \emph{NeurIPS}, 2017.

\bibitem[Wang et~al.(2021)Wang, Zhu, Adam, Yuille, and Chen]{wang2021max}
Huiyu Wang, Yukun Zhu, Hartwig Adam, Alan Yuille, and Liang-Chieh Chen.
\newblock Max-deeplab: End-to-end panoptic segmentation with mask transformers.
\newblock In \emph{CVPR}, 2021.

\bibitem[Wang et~al.(2023{\natexlab{a}})Wang, Zhang, Chu, Cao, Zhou, Wu, Wang, He, and Lin]{wang2023v3det}
Jiaqi Wang, Pan Zhang, Tao Chu, Yuhang Cao, Yujie Zhou, Tong Wu, Bin Wang, Conghui He, and Dahua Lin.
\newblock V3det: Vast vocabulary visual detection dataset.
\newblock In \emph{ICCV}, 2023{\natexlab{a}}.

\bibitem[Wang et~al.(2023{\natexlab{b}})Wang, Chen, Chen, Wu, Zhu, Zeng, Luo, Lu, Zhou, Qiao, and Dai]{wang2023visionllm}
Wenhai Wang, Zhe Chen, Xiaokang Chen, Jiannan Wu, Xizhou Zhu, Gang Zeng, Ping Luo, Tong Lu, Jie Zhou, Yu Qiao, and Jifeng Dai.
\newblock Visionllm: Large language model is also an open-ended decoder for vision-centric tasks.
\newblock \emph{NeurIPS}, 2023{\natexlab{b}}.

\bibitem[Wang et~al.(2023{\natexlab{c}})Wang, Dai, Chen, Huang, Li, Zhu, Hu, Lu, Lu, Li, Wang, and Qiao]{wang2023internimage}
Wenhai Wang, Jifeng Dai, Zhe Chen, Zhenhang Huang, Zhiqi Li, Xizhou Zhu, Xiaowei Hu, Tong Lu, Lewei Lu, Hongsheng Li, Xiaogang Wang, and Yu Qiao.
\newblock Internimage: Exploring large-scale vision foundation models with deformable convolutions.
\newblock In \emph{CVPR}, 2023{\natexlab{c}}.

\bibitem[Wang et~al.(2023{\natexlab{d}})Wang, Li, Chen, Lim, Torralba, Zhao, and Wang]{wang2023detecting}
Zhenyu Wang, Yali Li, Xi Chen, Ser-Nam Lim, Antonio Torralba, Hengshuang Zhao, and Shengjin Wang.
\newblock Detecting everything in the open world: Towards universal object detection.
\newblock In \emph{CVPR}, 2023{\natexlab{d}}.

\bibitem[Wei et~al.(2022{\natexlab{a}})Wei, Bosma, Zhao, Guu, Yu, Lester, Du, Dai, and Le]{wei2021finetuned}
Jason Wei, Maarten Bosma, Vincent~Y Zhao, Kelvin Guu, Adams~Wei Yu, Brian Lester, Nan Du, Andrew~M Dai, and Quoc~V Le.
\newblock Finetuned language models are zero-shot learners.
\newblock In \emph{ICLR}, 2022{\natexlab{a}}.

\bibitem[Wei et~al.(2022{\natexlab{b}})Wei, Wang, Schuurmans, Bosma, Xia, Chi, Le, and Zhou]{wei2022chain}
Jason Wei, Xuezhi Wang, Dale Schuurmans, Maarten Bosma, Fei Xia, Ed Chi, Quoc~V Le, and Denny Zhou.
\newblock Chain-of-thought prompting elicits reasoning in large language models.
\newblock \emph{NeurIPS}, 2022{\natexlab{b}}.

\bibitem[Xu et~al.(2023{\natexlab{a}})Xu, Liu, Vahdat, Byeon, Wang, and De~Mello]{xu2023open}
Jiarui Xu, Sifei Liu, Arash Vahdat, Wonmin Byeon, Xiaolong Wang, and Shalini De~Mello.
\newblock Open-vocabulary panoptic segmentation with text-to-image diffusion models.
\newblock In \emph{CVPR}, 2023{\natexlab{a}}.

\bibitem[Xu et~al.(2023{\natexlab{b}})Xu, Zhang, Wei, Hu, and Bai]{xu2023side}
Mengde Xu, Zheng Zhang, Fangyun Wei, Han Hu, and Xiang Bai.
\newblock Side adapter network for open-vocabulary semantic segmentation.
\newblock In \emph{CVPR}, 2023{\natexlab{b}}.

\bibitem[Yang et~al.(2023)Yang, Zhang, Li, Zou, Li, and Gao]{yang2023setofmark}
Jianwei Yang, Hao Zhang, Feng Li, Xueyan Zou, Chunyuan Li, and Jianfeng Gao.
\newblock Set-of-mark prompting unleashes extraordinary visual grounding in gpt-4v.
\newblock \emph{arXiv preprint arXiv:2310.11441}, 2023.

\bibitem[Yu et~al.(2020)Yu, Wang, Gao, Yu, Shen, and Sang]{yu2020context}
Changqian Yu, Jingbo Wang, Changxin Gao, Gang Yu, Chunhua Shen, and Nong Sang.
\newblock Context prior for scene segmentation.
\newblock In \emph{CVPR}, 2020.

\bibitem[Yu et~al.(2022{\natexlab{a}})Yu, Wang, Vasudevan, Yeung, Seyedhosseini, and Wu]{yu2022coca}
Jiahui Yu, Zirui Wang, Vijay Vasudevan, Legg Yeung, Mojtaba Seyedhosseini, and Yonghui Wu.
\newblock Coca: Contrastive captioners are image-text foundation models.
\newblock \emph{arXiv preprint arXiv:2205.01917}, 2022{\natexlab{a}}.

\bibitem[Yu et~al.(2022{\natexlab{b}})Yu, Wang, Kim, Qiao, Collins, Zhu, Adam, Yuille, and Chen]{yu2022cmt}
Qihang Yu, Huiyu Wang, Dahun Kim, Siyuan Qiao, Maxwell Collins, Yukun Zhu, Hartwig Adam, Alan Yuille, and Liang-Chieh Chen.
\newblock Cmt-deeplab: Clustering mask transformers for panoptic segmentation.
\newblock In \emph{CVPR}, 2022{\natexlab{b}}.

\bibitem[Yu et~al.(2022{\natexlab{c}})Yu, Wang, Qiao, Collins, Zhu, Adam, Yuille, and Chen]{yu2022k}
Qihang Yu, Huiyu Wang, Siyuan Qiao, Maxwell Collins, Yukun Zhu, Hartwig Adam, Alan Yuille, and Liang-Chieh Chen.
\newblock {k-means Mask Transformer}.
\newblock In \emph{ECCV}, 2022{\natexlab{c}}.

\bibitem[Yu et~al.(2023)Yu, He, Deng, Shen, and Chen]{yu2023convolutions}
Qihang Yu, Ju He, Xueqing Deng, Xiaohui Shen, and Liang-Chieh Chen.
\newblock Convolutions die hard: Open-vocabulary segmentation with single frozen convolutional clip.
\newblock \emph{NeurIPS}, 2023.

\bibitem[Zhang et~al.(2023)Zhang, Sun, Chen, Xiao, Shao, Zhang, Chen, and Luo]{zhang2023gpt4roi}
Shilong Zhang, Peize Sun, Shoufa Chen, Min Xiao, Wenqi Shao, Wenwei Zhang, Kai Chen, and Ping Luo.
\newblock Gpt4roi: Instruction tuning large language model on region-of-interest.
\newblock \emph{arXiv preprint arXiv:2307.03601}, 2023.

\bibitem[Zhao et~al.(2023)Zhao, Lin, Zhou, Huang, Feng, and Kang]{zhao2023bubogpt}
Yang Zhao, Zhijie Lin, Daquan Zhou, Zilong Huang, Jiashi Feng, and Bingyi Kang.
\newblock Bubogpt: Enabling visual grounding in multi-modal llms.
\newblock \emph{arXiv preprint arXiv:2307.08581}, 2023.

\bibitem[Zhou et~al.(2017)Zhou, Zhao, Puig, Fidler, Barriuso, and Torralba]{zhou2017scene}
Bolei Zhou, Hang Zhao, Xavier Puig, Sanja Fidler, Adela Barriuso, and Antonio Torralba.
\newblock Scene parsing through ade20k dataset.
\newblock In \emph{CVPR}, 2017.

\bibitem[Zhou et~al.(2023)Zhou, Liu, Yu, Li, Wang, and Wang]{zhou2023lmseg}
Qiang Zhou, Yuang Liu, Chaohui Yu, Jingliang Li, Zhibin Wang, and Fan Wang.
\newblock Lmseg: Language-guided multi-dataset segmentation.
\newblock In \emph{ICLR}, 2023.

\bibitem[Zhou et~al.(2022{\natexlab{a}})Zhou, Girdhar, Joulin, Kr{\"a}henb{\"u}hl, and Misra]{zhou2022detecting}
Xingyi Zhou, Rohit Girdhar, Armand Joulin, Philipp Kr{\"a}henb{\"u}hl, and Ishan Misra.
\newblock Detecting twenty-thousand classes using image-level supervision.
\newblock In \emph{ECCV}, 2022{\natexlab{a}}.

\bibitem[Zhou et~al.(2022{\natexlab{b}})Zhou, Koltun, and Kr{\"a}henb{\"u}hl]{zhou2022simple}
Xingyi Zhou, Vladlen Koltun, and Philipp Kr{\"a}henb{\"u}hl.
\newblock Simple multi-dataset detection.
\newblock In \emph{CVPR}, 2022{\natexlab{b}}.

\bibitem[Zhu et~al.(2023)Zhu, Chen, Shen, Li, and Elhoseiny]{zhu2023minigpt}
Deyao Zhu, Jun Chen, Xiaoqian Shen, Xiang Li, and Mohamed Elhoseiny.
\newblock Minigpt-4: Enhancing vision-language understanding with advanced large language models.
\newblock \emph{arXiv preprint arXiv:2304.10592}, 2023.

\bibitem[Zhu et~al.(2020)Zhu, Su, Lu, Li, Wang, and Dai]{zhu2020deformable}
Xizhou Zhu, Weijie Su, Lewei Lu, Bin Li, Xiaogang Wang, and Jifeng Dai.
\newblock Deformable detr: Deformable transformers for end-to-end object detection.
\newblock In \emph{ICLR}, 2020.

\end{thebibliography}
}

\clearpage
\setcounter{page}{1}
\maketitlesupplementary

In the supplementary materials, we provide more technical details of \modelname. We also include more
quantitative results and qualitative results, along with comparisons with GPT-4V~\cite{openai2023gpt}. Moreover, we show that \modelname can also be easily extended with part-level and box-level datasets, further unleasing the potential of \modelname.

\textbf{Instruction Template}\quad
We summarize the instruction template we used for \modelname training in \tabref{tab:instruction_template}.

\begin{figure}[t!]
    \centering
    \includegraphics[width=1.0\linewidth]{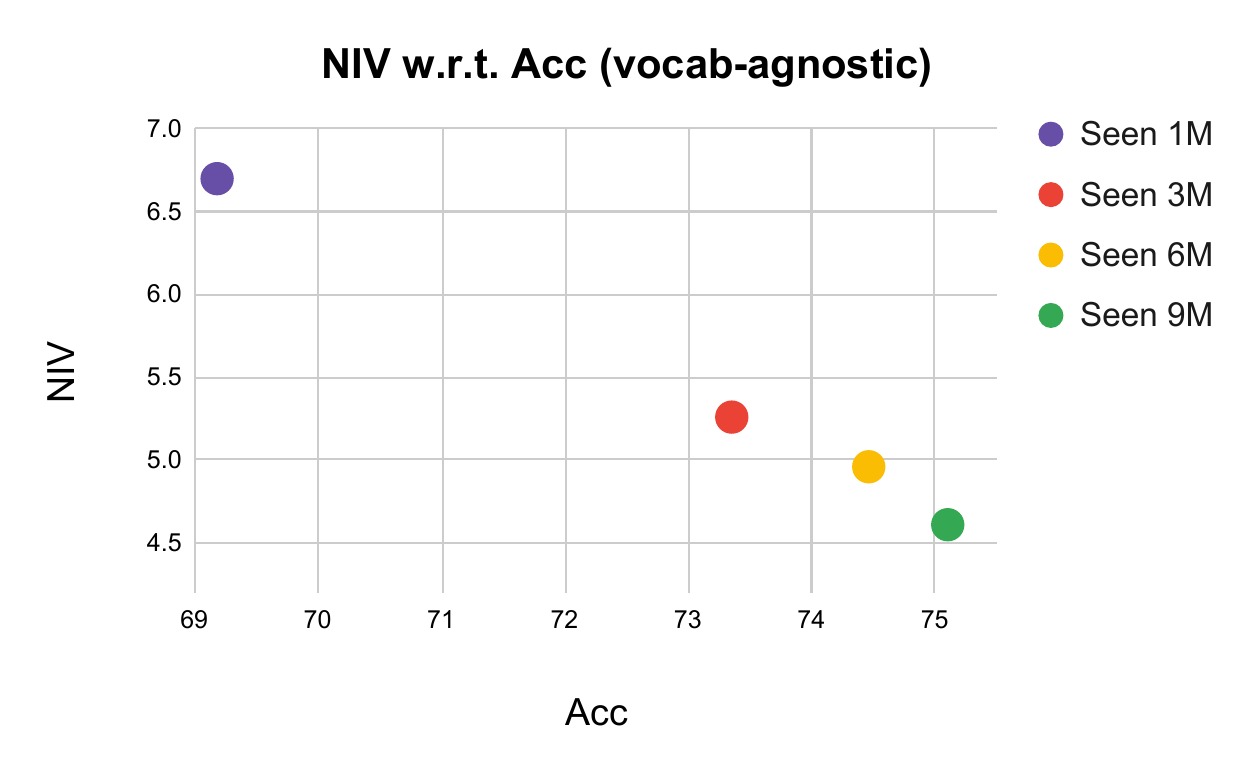}
    \caption{
    \textbf{NIV w.r.t. Acc, when number of seen masks varies.}
    The Acc increases as the number of seen masks increases, showing that the model is better trained to fit the target dataset. However, its NIV score becomes much lower, indicating that the model is losing the generalization ability. 
    }
    \label{fig:niv_wrt_acc}
\end{figure}

\textbf{Dilemma between Accuracy and Generalization}\quad
We also train \modelname under different seen number of masks (\ie, $1$, $3$, $6$, $9$ millions respectively), as shown in~\figref{fig:niv_wrt_acc}. Empirically, we consider Acc as a metric to measure how well the model can accurately recognize the object and NIV as a metric to measure the generalization ability. We note that there exists a dilemma between the accuracy and generalization, \ie, when the number of seen masks increases, we notice that the model achieves higher accuracy while inevitably having a higher overfitting to the training vocabulary, and predicts in a more conservative manner. From $6$M to $9$M, the accuracy improvement majorly comes from the decrease of NIV. We note that how to ensure an accurate object recognition while avoid overfitting to the training vocabulary is an interesting future research problem.

\begin{figure*}[t!]
    \centering
    \includegraphics[width=0.75\linewidth]{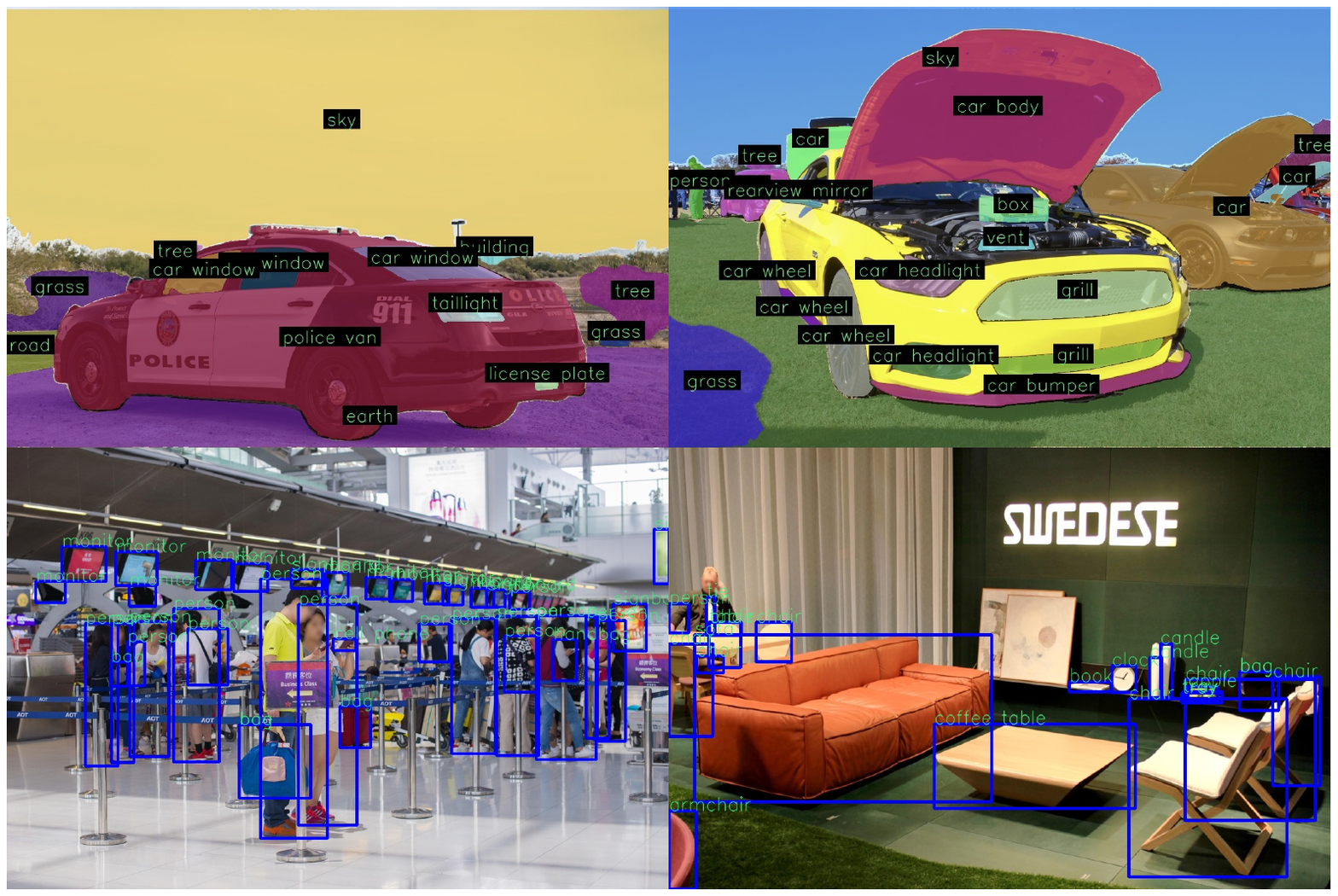}
    \vspace{-4ex}
    \caption{
    \textbf{Extending \modelname with part-level and box-level dataset.} We note that the \modelname framework is general and we can easily extend it with part-level and box-level data, leading to a stronger performance and more diverse usage. Best viewed zoom in to see predicted class names.
    }
    \label{fig:part_det_vis}
\end{figure*}

\textbf{Incorporating Part- and Box-level Datasets}\quad
It is worth noting that \modelname seamlessly accommodates part-level and box-level datasets, further enhancing its versatility. To enhance \modelname for part-level and box-level recognition (note that \modelname already shows emergent part recognition ability as illustrated in~\figref{fig:teaser_img}, but we believe introducing such datasets could further advance its ability), we introduce PartImageNet~\cite{he2022partimagenet}, Pascal-Part~\cite{chen2014detect}, and V3Det~\cite{wang2023v3det} datasets into the training data. For part data, we prepend the object name to part name, in case many parts sharing the same names (\eg, in PartImageNet, many different classes may have the same part named \textit{head}). We also remove those class names which are too vague (\eg, \textit{train left side}, \textit{bus upper side} in Pascal-Part). For detection data, we consider the bounding-box as a box-shaped binary mask and thus is easily unified into \modelname. Additionally, we augment the panoptic/instance segmentation data (\eg, COCO, LVIS) by randomly converting each segmentation mask into its corresponding bounding box. In cases where a bounding box serves as input, we appropriately adjust the instruction by replacing the term ``segmentation mask" with ``bounding box."
It's worth mentioning that we do not include image-level data (\eg, ImageNet) at this stage, as the semantic label could introduce bias when there exist multiple objects yet sharing single label. We demonstrate their effects in~\figref{fig:part_det_vis}, where we use SAM~\cite{kirillov2023sa-1b} and DETA~\cite{ouyang2022nms} as the proposal model respectively.

\begin{figure*}[th]
    \centering
    \includegraphics[width=0.9\linewidth]{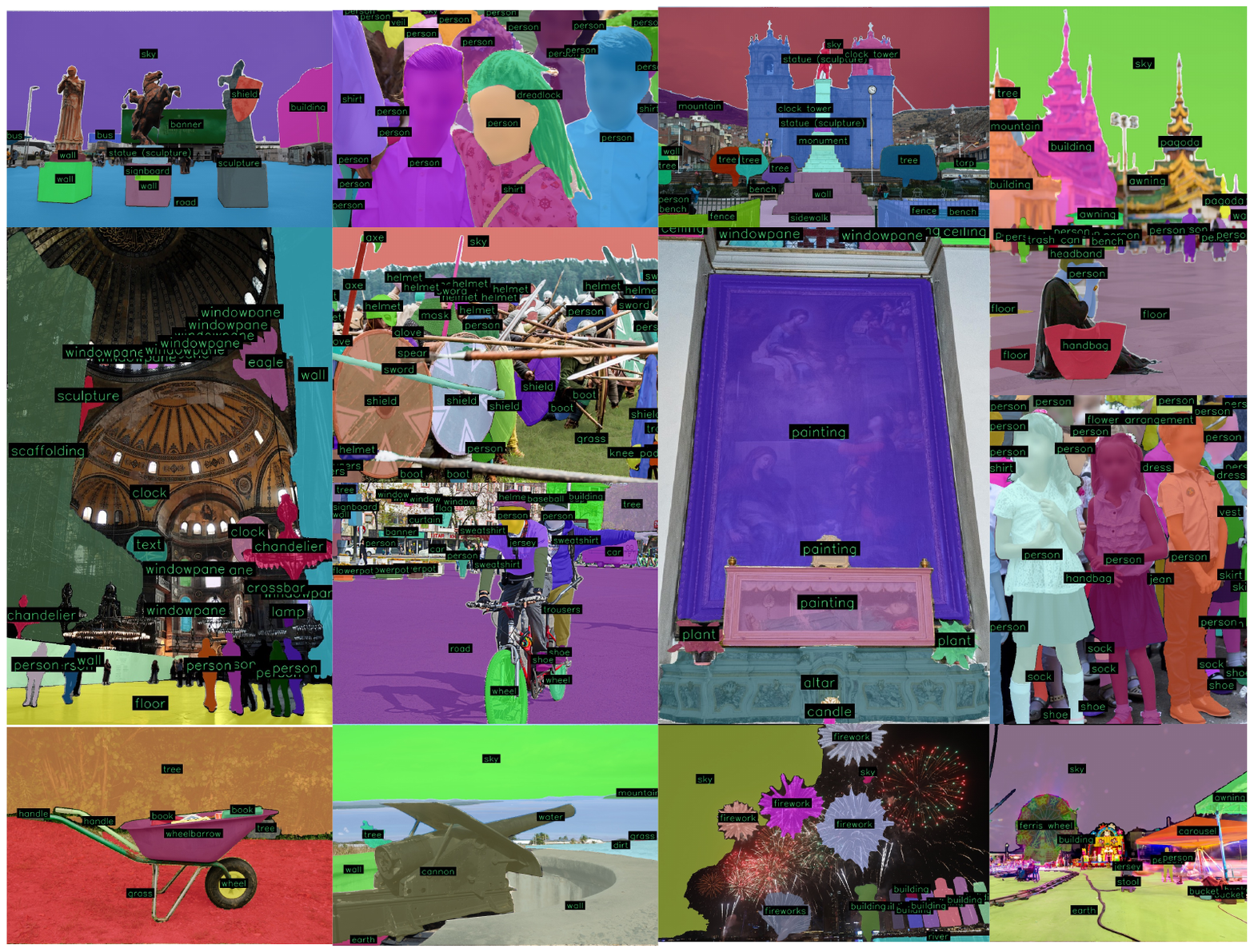}
    \vspace{-3ex}
    \caption{
    \textbf{Qualitative results on SA-1B dataset~\cite{kirillov2023sa-1b} of \modelname, using SAM as the mask proposal model.} Best viewed zoom in to see predicted class names.}
    \label{fig:vis}
\end{figure*}

\begin{figure*}[t]
    \centering
    \vspace{-10ex}
    \includegraphics[width=0.85\linewidth]{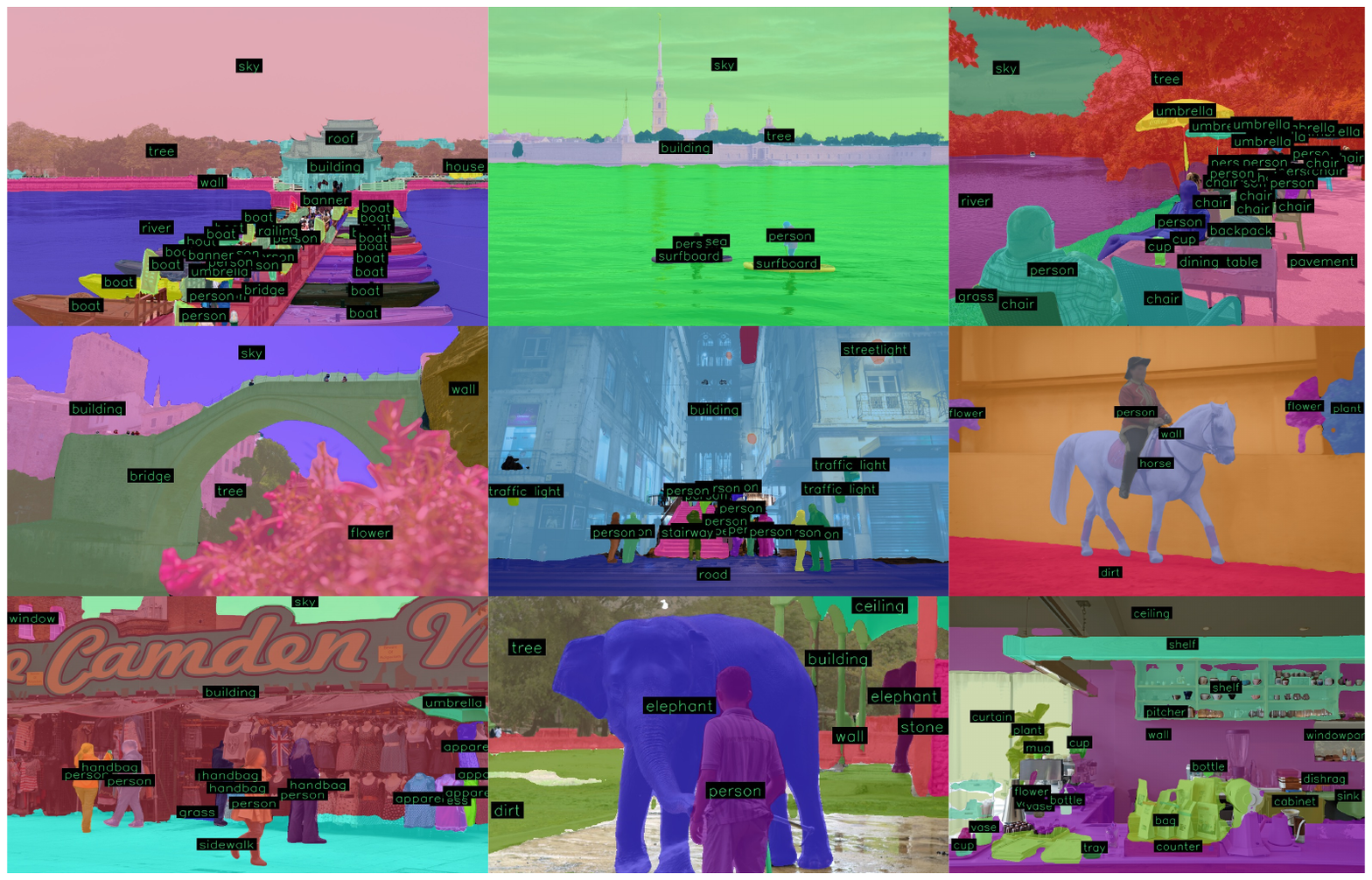}
    \vspace{-9ex}
    \caption{
    \textbf{Qualitative results on SA-1B dataset~\cite{kirillov2023sa-1b} of \modelname, using kMaX-DeepLab as the mask proposal model.} Best viewed zoom in to see predicted class names.}
    \label{fig:vis_kmax}
\end{figure*}

\textbf{Qualitative Results}\quad
We explore the application of \modelname on top of SAM~\cite{kirillov2023sa-1b} and kMaX-DeepLab~\cite{yu2022k}, and provide qualitative results, which are presented in~\figref{fig:vis} and~\figref{fig:vis_kmax} respectively. These results underscore the superiority of \modelname in practical scenarios and its potential to demonstrate open-ended recognition with fine-grained masks. When obtaining mask proposals from SAM~\cite{kirillov2023sa-1b}, we use the SAM variant with ViT-H~\cite{dosovitskiy2020image} backbone, with points per side $32$, IoU threshold $0.95$, stability threshold $0.95$, and minimum mask size $800$. This helps avoid too many small masks that are not recognizable (\eg, super-pixel level masks). When obtaining mask proposals from kMaX-DeepLab~\cite{yu2022k}, we use the one trained on COCO Panoptic dataset with ConvNeXt-L~\cite{liu2022convnet} backbone, and we set the ``thing" and ``stuff" threshold to $0.1$ to obtain more mask proposals and feed them into \modelname. Afterwards, we apply mask-wise post-processing following~\cite{yu2022cmt,yu2022k}.

\begin{table*}[t]
\centering
\tablestyle{2pt}{1.1}
\begin{tabular}{l}
\shline
1. What is in the segmentation mask? Assistant: \{class\_name\} \\
2. Describe what is in the segmentation mask. Assistant: \{class\_name\} \\
3. What does this segmentation mask show? Assistant: \{class\_name\} \\
4. What is this segmentation mask? Assistant: \{class\_name\} \\
5. What is the segmentation mask region of the image? Assistant: \{class\_name\} \\
6. Briefly describe what you perceive in the segmentation mask region. Assistant: \{class\_name\} \\
7. Please tell the category of what is indicated by the segmentation mask. Assistant: \{class\_name\} \\
8. What does this segmentation mask segments? Assistant: \{class\_name\} \\
9. What does this segmentation mask capture? Assistant: \{class\_name\} \\
10. Answer the name of what is in the segmentation mask region. Assistant: \{class\_name\} \\
11. What is the semantic class of the area given the segmentation mask? Assistant: \{class\_name\} \\
12. Can you describe what is in the segmentation mask region? Assistant: \{class\_name\} \\
13. From the image and segmentation mask provided, tell the category of the indicated region. Assistant: \{class\_name\} \\
14. Could you use a few words to describe what is in the segmentation mask region? Assistant: \{class\_name\} \\
15. Given the image and segmentation mask, answer what is in the region. Assistant: \{class\_name\} \\
16. Tell me what you see in the segmentation mask region. Assistant: \{class\_name\} \\
17. What can you see in the segmentation mask region? Assistant: \{class\_name\} \\
18. Let me know what you can perceive in the mask region. Assistant: \{class\_name\} \\
19. Give me the name of the object in the segmentation mask. Assistant: \{class\_name\} \\
\shline
\end{tabular}
\caption{
\textbf{Instruction templates.}
We randomly select one instruction template and insert the ground-truth class name during training. Only the first template \textit{What is in the segmentation mask?} is used during testing.
}
\label{tab:instruction_template}
\end{table*}

\begin{figure*}[t]
    \centering
    \vspace{-6ex}
    \includegraphics[width=0.9\linewidth]{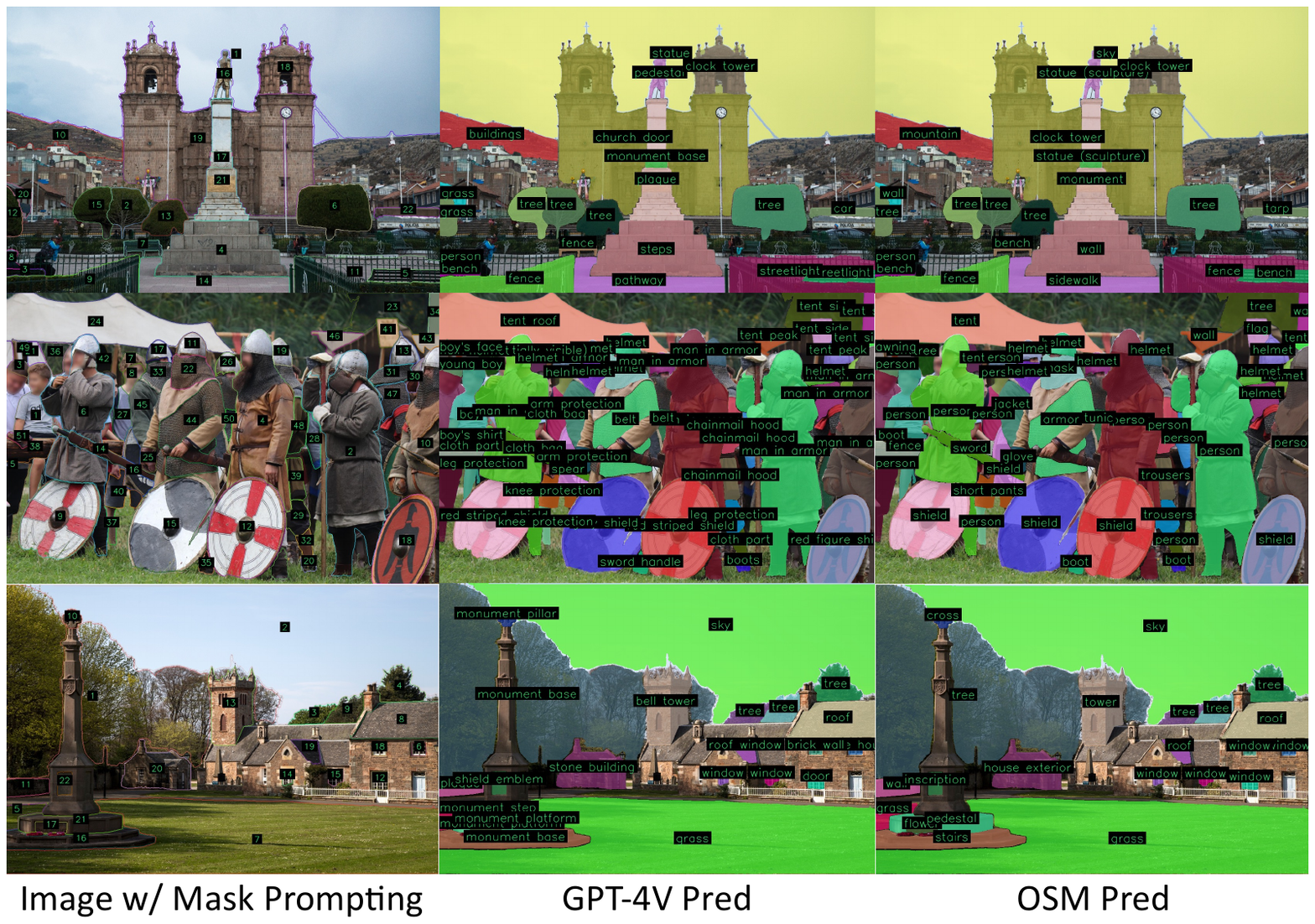}
    \vspace{-5ex}
    \caption{
    \textbf{Qualitative comparison \vs GPT-4V.} We follow~\cite{yang2023setofmark} to prompt GPT-4V for mask classification. Note that the enhanced image w/ mask prompting is only for GPT-4V input while \modelname still takes the original image as input. Best viewed zoom in to see predicted class names.}
    \label{fig:vis_comp_gpt4v}
\end{figure*}

\textbf{Comparison against GPT-4V}\quad
We provide a qualitative comparison between GPT-4V and \modelname. We follow~\cite{yang2023setofmark} to prompt GPT-4V for mask recognition. Specifically, we highlight the mask boundaries as auxiliary cues in the image, and annotate each mask center with a numeric ID. We feed the prompted image to GPT-4V, along with text prompt ``\textit{I have labeled a bright numeric ID at the center for each visual object in the image. Please enumerate their names (\ie, semantic class) with one, two, or three words.}". The results are shown in~\figref{fig:vis_comp_gpt4v}, with first column showing the image after mask prompting and fed to GPT-4V, and second column for GPT-4V predictions, third column for \modelname predictions. We observe that \modelname has a more accurate prediction compared to GPT-4V (\eg, in the first row, \modelname correctly predicts mask $5$ and $11$ as \textit{bench} and \textit{fence}, while GPT-4V wrongly predicts them both as \textit{streetlight}), which is often confused by the context (\eg, in the first row, for the mask $10$, GPT-4V predicts \textit{buildings} instead of \textit{mountain}, potentially due to confusion from the buildings below). However, we also note that \modelname's prediction is more conservative compared to GPT-4V, which can predict a more specific word. For example, in the second row, GPT-4V predicts \textit{man in armor} for the armed man in the image while \modelname still predicts in a safer way with \textit{person}. This also suggests a potential improvement of \modelname from a stronger base model (\eg, Llama2~\cite{touvron2023llama2}) or larger datasets with a better trade-off between accuracy and diversity~\cite{nguyen2023improving}. We also apply similar strategy to test with state-of-the-art open-sourced multi-modal large language model (\eg, LLava-1.5~\cite{liu2023improved}, MiniGPT-v2~\cite{chen2023minigptv2}) yet find them fail at generating reasonable outputs, coinciding observations in~\cite{yang2023setofmark}.

\begin{table}[t]
\centering
\tablestyle{10pt}{1.1}
\begin{tabular}{l|ccc}
methods & PQ & AP & mIoU  \\
\shline
OpenSeg~\cite{ghiasi2022scaling} & - & - & 21.1 \\
MaskCLIP~\cite{ding2022open} & 15.1 & 6.0 & 23.7 \\
FreeSeg~\cite{qin2023freeseg} & 16.3 & 6.5 & 24.6 \\
FC-CLIP~\cite{yu2023convolutions} & 17.8 & 11.1 & 20.8 \\
ODISE~\cite{xu2023open} & 19.5 & 10.8 & 23.8 \\
\hline
\modelname & 21.4 & 12.4 & 26.9  \\
\hline \hline
OVSeg$^*$~\cite{liang2023open} & - & - & 24.8 \\
SAN$^*$~\cite{xu2023side} & - & - & 33.3 \\
ODISE$^*$~\cite{xu2023open} & 23.4 & 13.9 & 28.7 \\
FC-CLIP$^*$~\cite{yu2023convolutions} & 26.8 & 16.8 & 34.1 \\
\hline
\modelname$^*$ & 26.9 & 16.2 & 33.6  \\
\end{tabular}
\caption{
\textbf{Comparison of \modelname in open-vocabulary settings on ADE20K \textit{val} set}. $*$: Methods using geometric ensemble from another frozen CLIP. We obtain ODISE and FC-CLIP's non-ensemble results by running their official code and setting  $\alpha=0$, $\beta=0$ in the Equation (7) of the FC-CLIP paper~\cite{yu2023convolutions}.
}
\vspace{-3ex}
\label{tab:openvocab_results}
\end{table}

\textbf{Evaluation with Open-Vocabulary Benchmarks}\quad
We also evaluate \modelname aginst state-of-the-art open-vocabulary methods in~\tabref{tab:openvocab_results}. To ensure an open-vocabulary setting (\ie, the target datasets are never seen during training), we train \modelname with COCO and LVIS data only and evaluate on ADE20K dataset in a zero-shot manner~\cite{ding2022open,xu2023open}. During testing, we use the same mask proposal model from~\cite{yu2023convolutions}, and replace the classification head with \modelname. Furthermore, we map \modelname's prediction to target vocabulary using text embedding similarity between predicted class name and target vocabulary class names. Following prior arts~\cite{xu2023open,yu2023convolutions}, we also apply geometric ensemble to enhance the results with the frozen CLIP predictions.
We report results with and without ensemble.
As shown in the table, when not using the geometric ensemble method, \modelname shows superior scores against state-of-the-art open-vocabulary methods, indicating its strong performance.
When using the geometric ensemble from another frozen CLIP, \modelname still shows a comparable performance with other state-of-the-art methods, such as ODISE~\cite{xu2023open} and FC-CLIP~\cite{yu2023convolutions}.

\begin{figure*}[t]
    \centering
    \vspace{-9ex}
    \includegraphics[width=0.9\linewidth]{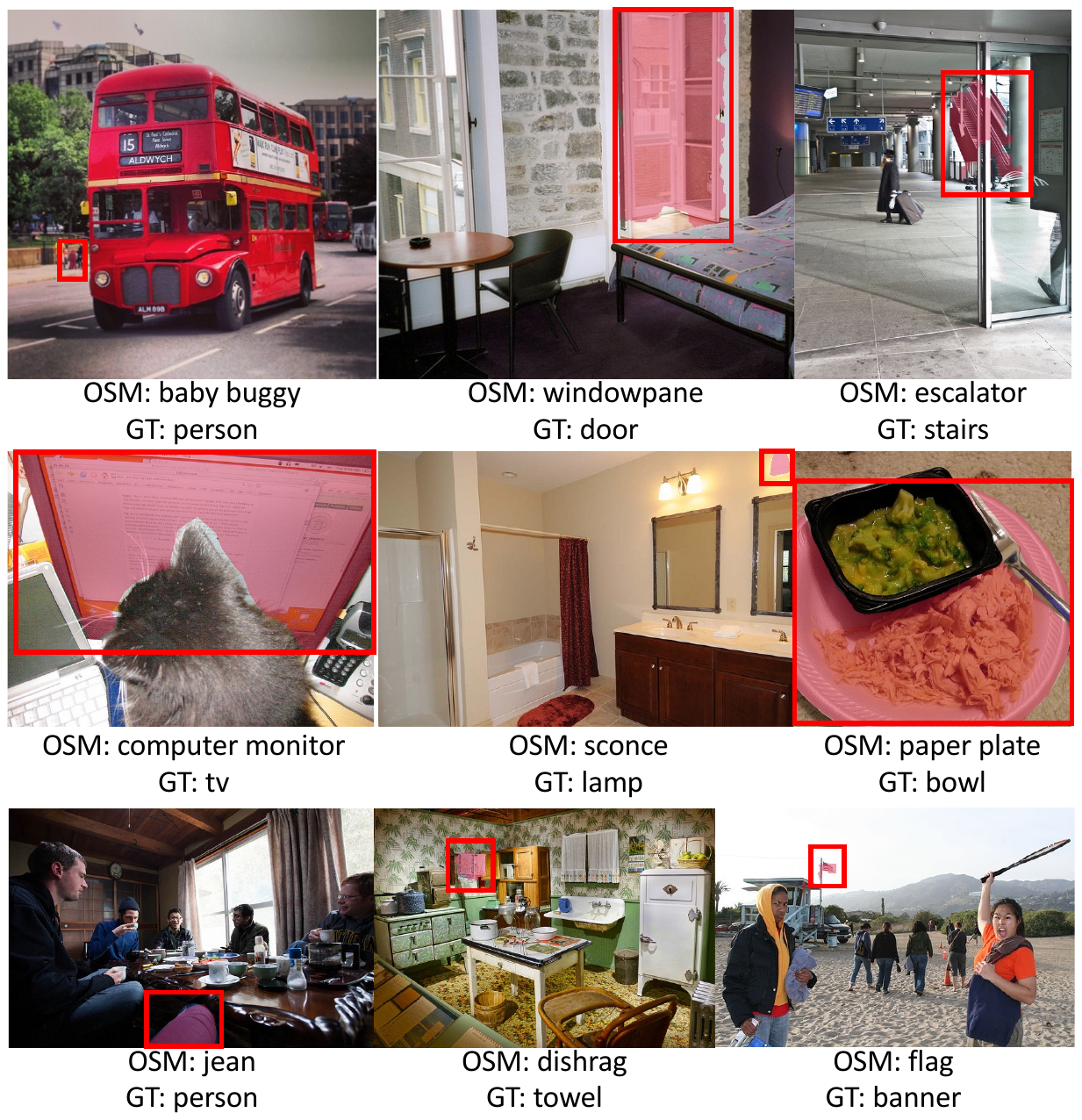}
    \vspace{-15ex}
    \caption{
    \textbf{Visualization of NIV cases on COCO \textit{val} set using ground-truth mask}. We note that \modelname can predict a more felicitous class compared to ground-truth, where annotations are limited to the fixed vocabulary and thus usually less expressive. We highlight the mask region with bounding box for better visualization purposes. Best viewed zoom in.}
    \label{fig:vis_niv_coco}
\end{figure*}

\begin{figure*}[t]
    \centering
    \vspace{-9ex}
    \includegraphics[width=0.9\linewidth]{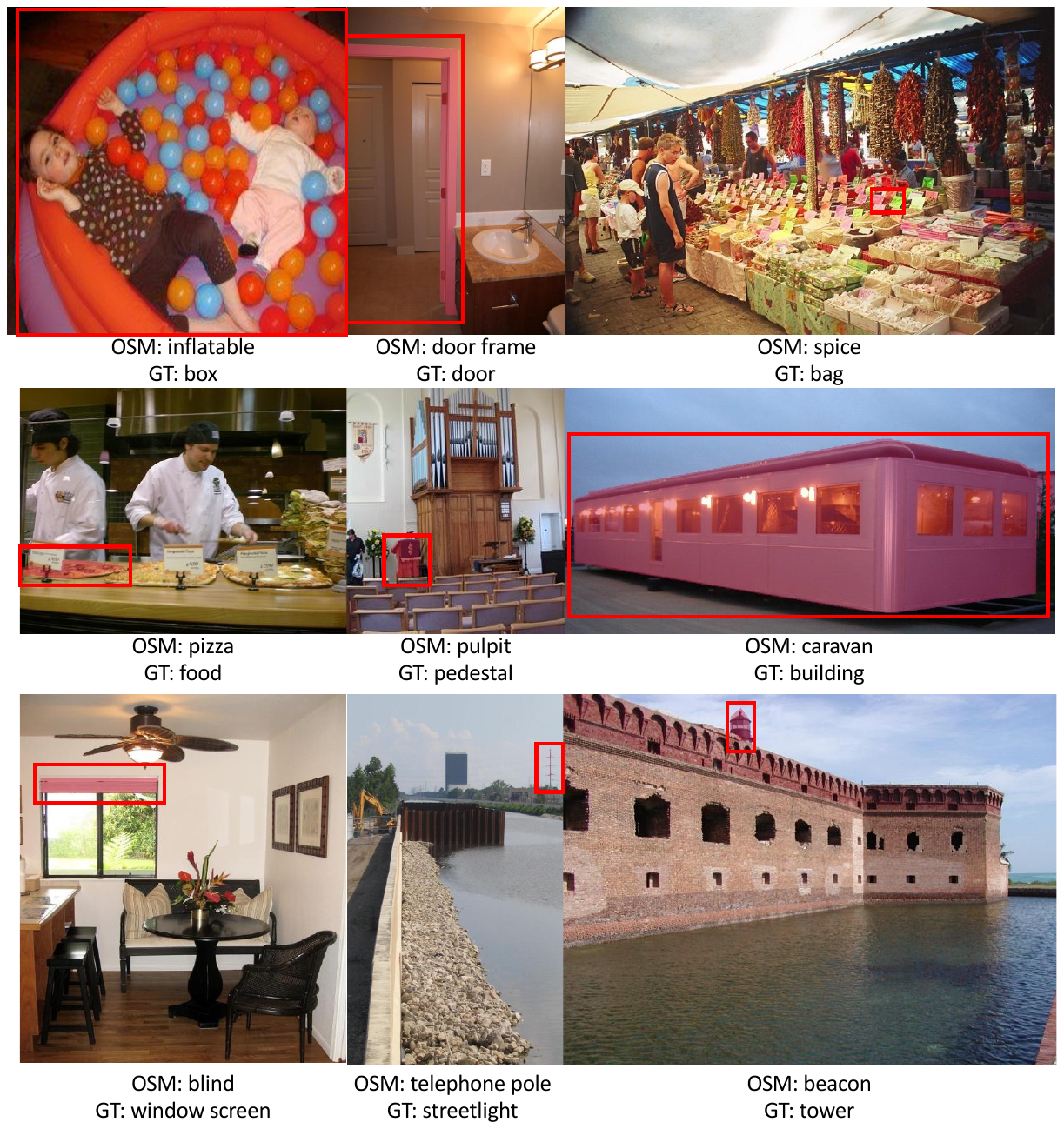}
    \vspace{-15ex}
    \caption{
    \textbf{Visualization of NIV cases on ADE20K \textit{val} set using ground-truth mask}. We note that \modelname can predict a more felicitous class compared to ground-truth, where annotations are limited to the fixed vocabulary and thus usually less expressive. We highlight the mask region with bounding box for better visualization purposes. Best viewed zoom in.}
    \label{fig:vis_niv_ade20k}
\end{figure*}

\textbf{Visualization of NIV Cases}\quad
To better understand what are NIV (Not-in-Vocab) cases, we visualize them using ground-truth mask against ground-truth annotation for COCO \textit{val} set and ADE20K \textit{val} set in~\figref{fig:vis_niv_coco} and \figref{fig:vis_niv_ade20k} respectively. We note that with a pre-defined vocabulary, even the ground-truth annotations are usually biased and limited, where annotators have to pick a most similar class in the given vocabulary (\eg, all \textit{monitor} are labeled as \textit{tv} in COCO). The biases could be learnt and inherited in existing closed-vocabulary and open-vocabulary models. However, we observe the \modelname can predict a more appropriate class name without limitation of a given vocabulary, demonstrating the necessity and effectiveness of getting rid of a pre-defined vocabulary and pursuing open-ended visual recognition.

\end{document}